\documentclass{article} %
\usepackage{iclr2027_conference,times}

\usepackage{amsmath,amsfonts,bm}

\def\eqref#1{equation~\ref{#1}}

\def\1{\bm{1}}

\DeclareMathAlphabet{\mathsfit}{\encodingdefault}{\sfdefault}{m}{sl}
\SetMathAlphabet{\mathsfit}{bold}{\encodingdefault}{\sfdefault}{bx}{n}

\usepackage{hyperref}
\usepackage{url}
\usepackage{capt-of}
\usepackage{xcolor}
\usepackage{booktabs}
\usepackage{amssymb}
\usepackage{graphicx}
\usepackage{afterpage}
\usepackage{longtable}
\usepackage{array}
\usepackage{colortbl}
\usepackage{wrapfig}
\usepackage{caption}
\usepackage{cleveref}
\usepackage{enumitem}

\title{PrismWAM: Decoupling State Representations in World Action Models}

\author{%
  \mbox{Tianyu Xin\textsuperscript{1,*}}, \mbox{Wanqi Zhong\textsuperscript{1,*}},
  \mbox{Bowen Hu\textsuperscript{1}}, \mbox{Zhangchi Guo\textsuperscript{1}},
  \mbox{Lijie Wang\textsuperscript{1}},\\
  \mbox{Yiming Wang\textsuperscript{1}}, \mbox{Haoran Liao\textsuperscript{1}},
  \mbox{Heyang Yu\textsuperscript{1}}, \mbox{Xiaoyu Tian\textsuperscript{1}},
  \mbox{Zhiding Yu\textsuperscript{2}}, \mbox{Yiming Li\textsuperscript{1,\ensuremath{\dagger}}}%
}

\makeatletter
\renewcommand{\@maketitle}{%
  \vbox{\hsize\textwidth\parindent\z@\raggedright
    \vskip -58pt
    \hbox to\textwidth{%
      \includegraphics[height=32.4pt]{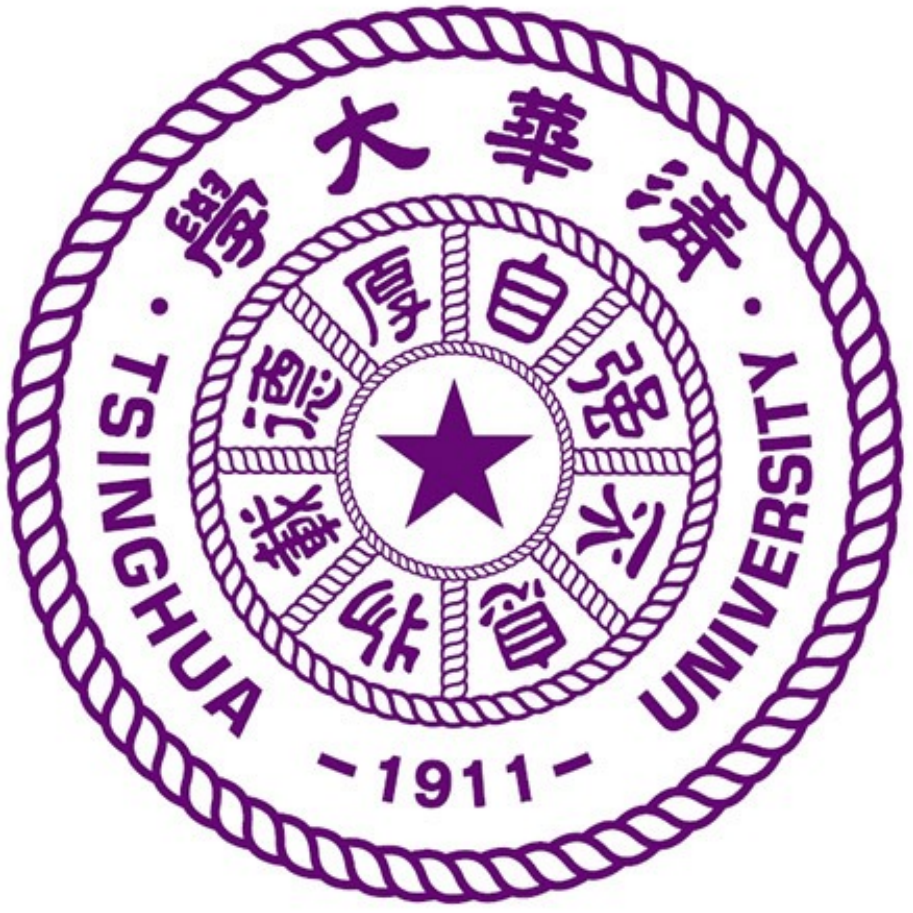}%
      \hspace{16.2pt}%
      \raisebox{7.2pt}{\includegraphics[height=18pt]{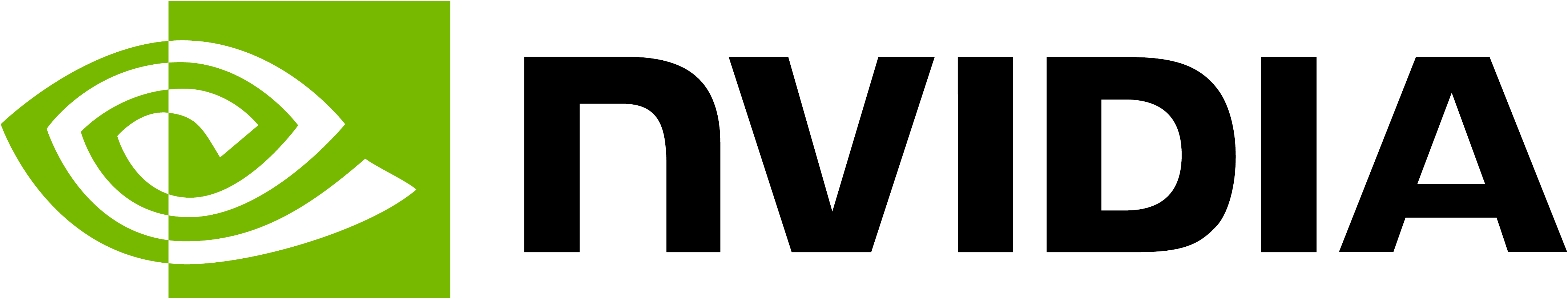}}%
      \hfil}%
    \vskip 7.5pt
    {\color[gray]{0.55}\hrule height 0.4pt}%
    \vskip 18pt
    {\fontsize{16}{20}\selectfont\scshape
      \hangindent=1.5em\hangafter=1
      \noindent\raisebox{-0.25em}[0pt][0pt]{\includegraphics[width=1.15em]{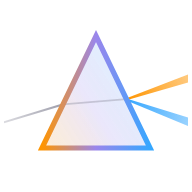}}%
      \hspace{0.35em}\@title\par}%
    \vskip 12pt
    {\fontsize{10.5}{13.5}\selectfont\bfseries\@author\par}%
    \vskip 6pt
    {\fontsize{9.5}{12}\selectfont
      \textsuperscript{1}Tsinghua University\quad
      \textsuperscript{2}NVIDIA\\
      \textsuperscript{*}Equal contribution\quad
      \textsuperscript{\ensuremath{\dagger}}Corresponding author\par}%
    \vskip 4pt
    {\small Project Page: \url{https://prism-wam.github.io/}\par}%
    \vskip 10pt
  }%
}
\makeatother

\iclrfinalcopy %
\fancypagestyle{prismfirstpage}{\fancyhead{}\renewcommand{\headrulewidth}{0pt}}

\begin{document}
\thispagestyle{prismfirstpage}

\afterpage{%
    \begin{figure}[!t]
        \centering
        \includegraphics[width=0.98\textwidth]{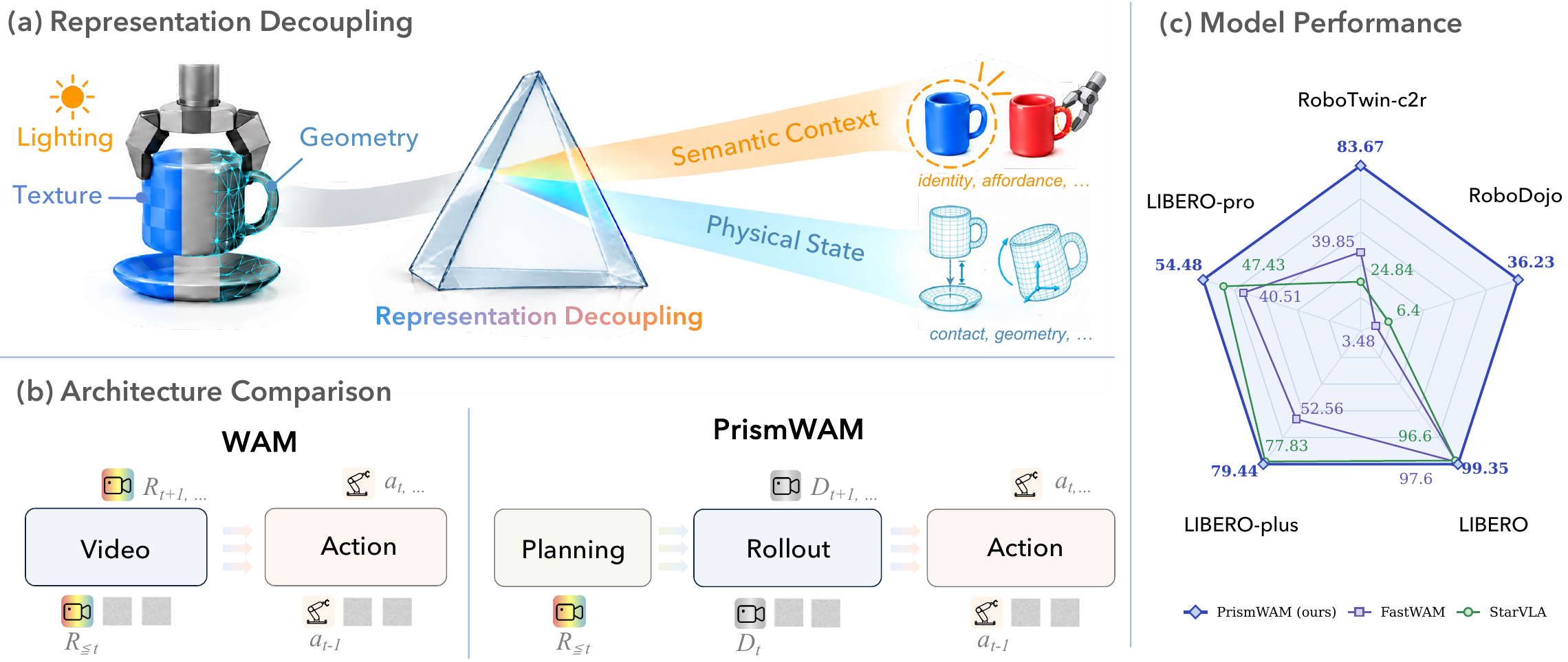}
        \caption{\textbf{Overview of PrismWAM.} (a) Our key insight is to decouple physical state from semantic context, yielding state representations that are better aligned with action-induced dynamics. (b) We follow a causal workflow of \textit{what to do} $\rightarrow$ \textit{how the world responds} $\rightarrow$ \textit{how to act}, corresponding to \textit{Semantic Planning} $\rightarrow$ \textit{Physical Rollout} $\rightarrow$ \textit{Executable Action}. (c) Comparison with starVLA and FastWAM across five benchmarks.}
        \label{fig:teaser}
    \end{figure}%
}

\maketitle

\begin{abstract}
World Action Models (WAMs) typically represent the world through latent embeddings from pretrained video models, which entangle heterogeneous aspects of the world, from \textit{how it looks} (e.g., texture and color) to \textit{how it is physically configured} (e.g., occupancy and pose). This raises an underexplored question: \textbf{\textit{what should constitute the state of a WAM?}} We argue that the world state should separate \textit{how the world appears} from \textit{how the world evolves under action}: the former captures \textit{semantic context}, while the latter is governed by \textit{physical state}. To this end, we propose \textbf{PrismWAM}, which decouples \textit{physics} from \textit{semantics} while jointly modeling state and action. PrismWAM factorizes an embodied task into a $\mathrm{Semantic\ Planning}\rightarrow\mathrm{Physical\ Rollout}\rightarrow\mathrm{Executable\ Action}$ pipeline: given only the initial RGB observation, semantic planning learns to predict the final goal image, while physical rollout predicts the fine-grained evolution toward this goal in depth space; executable actions are then generated from the depth rollout. Without any embodied action pretraining or additional data augmentation, PrismWAM achieves strong manipulation performance in both simulation and the real world. Notably, PrismWAM even surpasses a broad range of models with large-scale embodied action pretraining (e.g., OpenWAM-$\alpha$). More broadly, we reframe World Action Models from modeling \textit{how the world looks} to \textit{how it responds to action}, paving the way for action-centric world models grounded in occupancy, flow, tactile sensing, proprioception, and beyond.
\end{abstract}

\section{Introduction}
World models have emerged as a promising paradigm for enabling autonomous agents to predict the consequences of their actions~\citep{ha2018world,lecun2022path}. By learning how the world evolves over time, these models aim to provide a foundation for planning and action~\citep{hafner2019dream}. Recent World Action Models (WAMs) extend this paradigm by jointly modeling video and action, often relying on pretrained video backbones to capture the rich visual dynamics of the physical world~\citep{dreamzero,lingbot-va,gigaworld-policy}. Yet, a key question remains underexplored: \textbf{\textit{are video latents truly the most appropriate representations for world states?}}

Our key observation is that video latents capture a mixture of factors that are not equally relevant to action generation~\citep{cosmos-predict2.5}. While appearance and texture provide rich semantic context for understanding the scene, they are often invariant to the actions being modeled. For example, when an agent moves or grasps an object, its geometry and pose may change, while its color, texture, and intrinsic appearance typically remain unchanged. Therefore, video latents which entangles the \textbf{physical state} with the \textbf{semantic context} can make the representation substantially richer than what is necessary for modeling how the world evolves under action.

This motivates a simple principle: a good state representation in WAMs should distinguish between {physical state} and {semantic context}. The physical state refers to action-relevant properties such as geometry, pose, occupancy, contact, and potentially dynamics, whereas semantic context provides the appearance information necessary to interpret that state, such as object identity, category, and texture; see~Fig.~\ref{fig:teaser} (a). The distinction is not between information that is useful and useless; rather, it decouples {\textit{what actions can change}} from {\textit{what helps us understand what is being acted upon}}.

Therefore, we propose \textbf{PrismWAM} to separate semantic context from physical dynamics through a $\mathrm{Semantic~Planning}\rightarrow\mathrm{Physical~Rollout}\rightarrow\mathrm{Executable~Action}$ workflow (see~Fig.~\ref{fig:teaser} (b)), rather than treating a single video latent as a monolithic state. Semantic planning infers \textbf{\textit{what should happen in the end}} from RGB observations, physical rollout models \textbf{\textit{how the world physically evolves}} toward the goal, and executable action determines \textbf{\textit{how to make it happen}}. We instantiate this principle with a Mixture-of-Transformers (MoT) architecture built upon a pretrained video backbone~\citep{wan2.2}, as shown in Fig.~\ref{fig:method_architecture}. During training, the semantic planner predicts the target from the initial RGB observation and transfers its task-level knowledge to the physical rollout via attention layers, while the action branch learns to map the predicted physical evolution to continuous actions. We use depth as the physical representation for its accessibility, physical grounding, and spatial alignment with RGB. Notably, the semantic planner serves only as a training-time teacher, with action inference relying solely on the initial RGB observation and learned depth rollout.

Extensive experiments across diverse simulation and real-world benchmarks validate the effectiveness and efficiency of PrismWAM. \textbf{(1)}~Without action pretraining, PrismWAM substantially outperforms representative VLAs and WAMs under the same setting, improving the success rate over FastWAM by \textbf{43.82} percentage points ($83.67$ vs. $39.85$) on RoboTwin-C2R and the score over starVLA by \textbf{29.83} points ($36.23$ vs. $6.40$) on RoboDojo; see Fig.~\ref{fig:teaser}(c). \textbf{(2)}~PrismWAM outperforms several WAMs, including models with action pretraining, with particularly strong gains in precision manipulation and generalization: on RoboDojo Precision, it surpasses OpenWAM-$\alpha$ by \textbf{21.37} points ($39.82$ vs. $18.45$), while improving the success rate on RoboTwin-C2R by \textbf{14.62} percentage points ($83.67$ vs. $69.05$), showing a higher ceiling for complex tasks and novel scenes. \textbf{(3)}~PrismWAM even matches or surpasses large-scale pretrained models on several benchmarks despite being trained with only $\sim$2,300 hours of depth data. On RoboDojo, it improves the score by \textbf{16.00} points over G0.5 ($36.23$ vs. $20.23$) and \textbf{16.16} points over Xiaomi-Robotics-1 ($36.23$ vs. $20.07$), showing that scaling action pretraining is not the only path to stronger embodied foundation models; principled architectural design offers a more data-efficient alternative. Our key contributions are listed as follows:
\begin{itemize}[leftmargin=6mm]
\item \textbf{A decoupling principle for World Action Models.} We identify the entanglement of semantics and physics in video latents as a fundamental limitation of existing WAMs, forcing action generation to disentangle high-level task planning from fine-grained physical state transitions.

\item \textbf{An effective and efficient WAM architecture.} We instantiate the principle with a Mixture-of-Transformers (MoT), where an RGB semantic planner guides depth-based physical rollout, and develop a learning pipeline combining dense depth supervision with sparse RGB goal supervision, enabling fast inference without explicit RGB goal prediction.

\item \textbf{Strong performance and data efficiency.} PrismWAM outperforms existing VLAs and WAMs, with particularly strong gains in precision manipulation and generalization. It also surpasses a number of models with large-scale action pretraining, demonstrating that principled architectural design offers a more data-efficient path toward stronger embodied foundation models.
\end{itemize}
\section{Related Works}

\noindent\textbf{Vision-Language-Action Policies.}
The general and embodied knowledge encoded in pretrained vision-language models (VLMs)~\citep{gemini,qwen-vl,rynnbrain,hy-embodied,mimo-embodied} provides a foundation for vision-language-action (VLA) models that map visual observations and language instructions to robot actions~\citep{RT-2,openvla,pi_0,cogact,pi05,gr00t-n1,smolvla}. Internet-scale image--text pretraining provides semantic priors for complex instruction understanding and long-horizon manipulation~\citep{RT-2,pi_0,pi05,gr00t-n1,hi-robot}. However, such pretraining emphasizes recognition and semantic alignment, with limited explicit supervision for continuous state transitions and how actions change the environment.

\noindent\textbf{World Action Models.}
Advances in video generative models~\citep{wan2.2,cosmos-predict2.5,cosmos3} have paved the way for world action models (WAMs), a promising paradigm that jointly models future states and robot actions for generalist robotic manipulation. Most jointly model future RGB observations and action sequences~\citep{uwm,cosmos-policy}. Some methods guide actions based on predicted future RGB observations~\citep{videovla,motus,lingbot-va,dreamzero,next-forcing,lingbot-va2,openwam}. Others use future prediction as training supervision and generate actions directly from observed context~\citep{fastwam,gigaworld-policy,pfd,arps,meco-wam}. Both approaches primarily learn state evolution in RGB appearance space, which entangles action-relevant geometry and pose with texture, illumination, and background. This requires modeling appearance details that may be irrelevant to control and inferring physical changes from visual variations that may not reflect motion~\citep{agra,meco-wam}.

\noindent\textbf{Physical Representations for Manipulation.}
Physical or spatial representations support precise manipulation and spatial generalization~\citep{peract,rvt,dp3,3d-diffuser-actor,spatialvla}. Such information can directly guide action generation through depth or point-cloud features extracted from current observations~\citep{depthvla,pointvla,3d-diffuser-actor}. Complementary approaches introduce physical or spatial structure through supervision, including representation alignment~\citep{spatial-forcing,lingbot-vla2,3dthinkvla,4d-wam} and auxiliary prediction~\citep{xwam,wam4d,meco-wam,dreamvla}.

Recent world action models take RGB and geometry as joint inputs and predict their future states together~\citep{rynnworld-4d,sa-wam}. This couples semantic and physical information, retaining the RGB-based modeling issues discussed above. PrismWAM separates their roles along a $\mathrm{Semantic\ Planning}\rightarrow\mathrm{Physical\ Rollout}\rightarrow\mathrm{Executable\ Action}$ chain, distinguishing semantic task understanding from the physical state evolution that guides action execution.

\section{Method}
\label{sec:method}

PrismWAM follows a $\mathrm{Semantic\ Planning}\rightarrow\mathrm{Physical\ Rollout}\rightarrow\mathrm{Executable\ Action}$ chain for prediction and control. The RGB stream learns semantic planning representations from a sparse future-frame target, while the depth stream predicts a dense physical rollout that guides action generation. We first present the problem formulation (\S\ref{sec:method_formulation}), then describe PrismWAM's architecture, training, and inference (\S\ref{sec:method_architecture}).

\begin{figure}[t]
    \centering
    \includegraphics[width=0.98\linewidth]{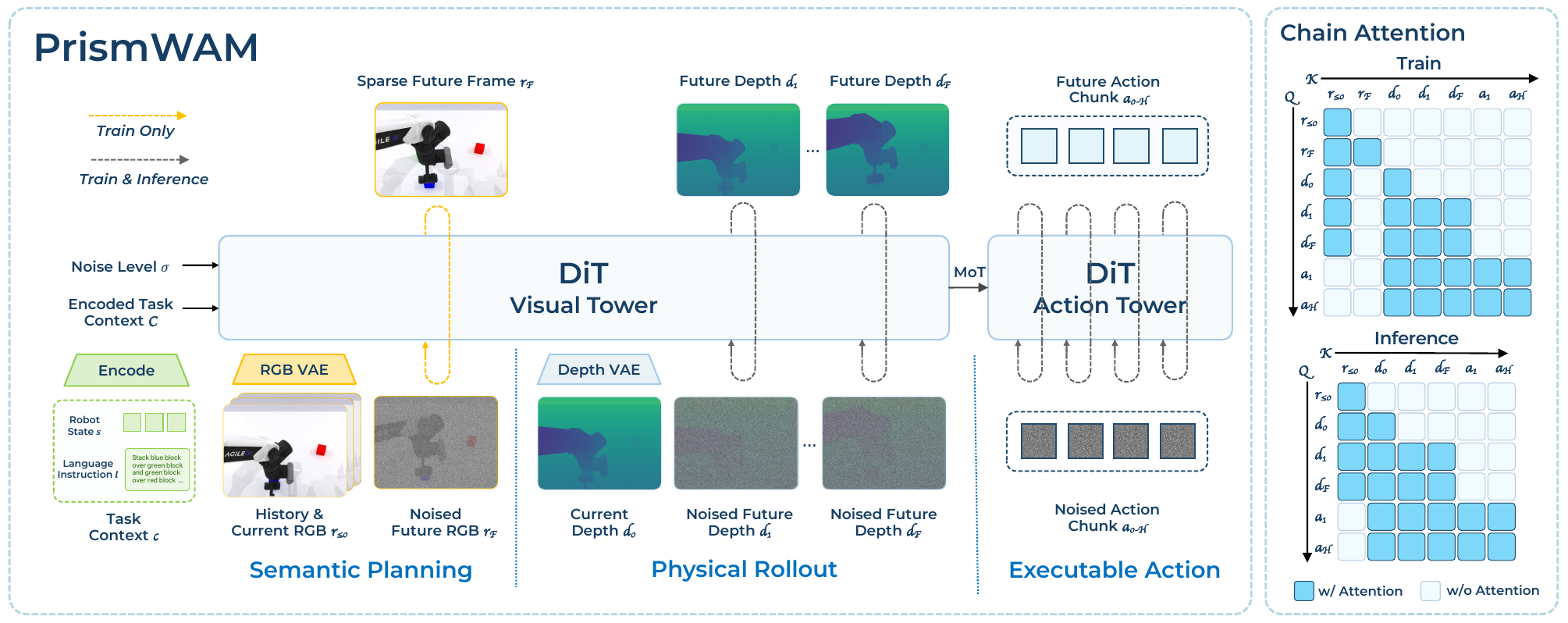}
    \caption{\textbf{PrismWAM decouples semantic planning from physical rollout.} \textbf{Left}: the RGB stream learns task semantics through sparse supervision from a single intended-future frame, while the depth stream predicts a dense geometric trajectory jointly with actions. \textbf{Right}: chain attention routes observed-RGB features to the depth stream and physical rollout features to the action tower.}
    \label{fig:method_architecture}
    \vspace{-3mm}
\end{figure}

\subsection{Problem Formulation}
\label{sec:method_formulation}

We formulate a world--action model (WAM) that jointly predicts world-state evolution and the corresponding actions. Let $x_0$ denote the current world state, $x_{1:F}$ its future world state trajectory, and $a_{1:H}$ the corresponding action sequence. Given task context $c=(\ell,s)$, comprising the language instruction $\ell$ and proprioceptive state $s$, the policy can be expressed as
\begin{equation}
    \pi_\theta(a_{1:H}\mid x_0,c)
    =\int
    \underbrace{p_\theta(a_{1:H}\mid x_{1:F},x_0,c)}_{\text{action generation}}
    \underbrace{p_\theta(x_{1:F}\mid x_0,c)}_{\text{world-state evolution}}
    \,\mathrm{d}x_{1:F}.
\label{eq:world_action_formulation}
\end{equation}
Here, $p_\theta(x_{1:F}\mid x_0,c)$ models how the current state may evolve over the prediction horizon, while $p_\theta(a_{1:H}\mid x_{1:F},x_0,c)$ predicts actions conditioned on the current state and a future state trajectory. Marginalizing over possible future world state trajectories yields the policy's action distribution. The future world state trajectory $x_{1:F}$ and action sequence $a_{1:H}$ cover the same time interval.

\paragraph{Semantic Planning and Physical Rollout.}
We use historical and current RGB observations $r_{\leq 0}$ for semantic planning and the current depth observation $d_0$ as the initial physical state. Semantic planning captures task intent and the intended scene configuration, while physical rollout models the intermediate geometric changes needed for action execution. Guided by RGB representations, the depth stream predicts the sequence $d_{1:F}$, and the action tower jointly generates executable actions from the evolving rollout features.

Let $z_R$ denote semantic planning representations extracted from $r_{\leq 0}$ and $z_D$ the physical rollout features produced during depth prediction. The resulting policy is
\begin{equation}
    \pi_\theta(a_{1:H}\mid r_{\leq 0},d_0,c)
    =\iint
    \underbrace{p_\theta(a_{1:H}\mid z_D,c)}_{\text{Executable Action}}\,
    \underbrace{p_\theta(z_D\mid d_0,z_R,c)}_{\text{Physical Rollout}}\,
    \underbrace{p_\theta(z_R\mid r_{\leq 0},c)}_{\text{Semantic Planning}}
    \,\mathrm{d}z_R\,\mathrm{d}z_D.
\label{eq:plan_transition_action}
\end{equation}

\subsection{PrismWAM}
\label{sec:method_representation}
\label{sec: arch}
\label{sec:method_architecture}

\paragraph{Architecture with Sparse--Dense Temporal Supervision.}
Building on the pretrained video backbone WAN~\citep{wan2.2}, PrismWAM uses RGB and depth streams with asymmetric temporal supervision, coupled to an action tower through a mixture-of-transformers (MoT) architecture (Fig.~\ref{fig:method_architecture}, left). The RGB stream predicts a single intended-future frame $r_F$, a sparse target for learning task semantics. The depth stream predicts the denser sequence $d_{1:F}$, providing intermediate spatial states that guide action evolution. This assigns semantic planning to sparse RGB prediction and physical rollout to dense geometric prediction, whose representation is less entangled with texture and illumination.

The model receives RGB history $r_{\leq 0}$, current depth $d_0$, instruction $\ell$, and proprioception $s$. Camera views are resized into matching RGB and depth mosaics, and RGB values and clipped metric depth are normalized. The frozen RGB and depth VAEs encode observations and future training targets into latent representations, which modality-specific patch projections convert into visual tokens. Frozen T5 embeddings and projected, normalized proprioception form the context $C$, which all branches access through cross-attention. Appendix~\ref{appn:implementation} specifies input layouts and temporal sampling.

During joint training, the RGB token sequence combines clean history latents with noisy latents of the single future target, while the depth sequence combines clean current-depth latents with noisy future-depth latents. The action branch linearly projects each action vector in the noisy sequence into an action token. Gaussian corruption is applied only to prediction targets; observed RGB and depth remain fixed. Each branch receives its own noise-level embedding.

\paragraph{Chain Attention.}
A structured attention mask controls information flow between these token groups (Fig.~\ref{fig:method_architecture}, right). Within each visual stream, future tokens attend to observed tokens and one another, while observed tokens cannot read future tokens. Depth tokens additionally read only the observed-RGB tokens. Action tokens attend to one another and to both observed and future depth tokens, with no direct access to RGB. These cross-modal connections are directed: RGB does not read depth or actions, and depth does not read actions. Thus, action prediction uses evolving depth features without accessing the future-RGB target.

Specifically, let $h_m^{(j)}$, $m\in\{R,D,A\}$, denote the RGB, depth, and action features after $j$ layers, with $h_m^{(0)}$ their input embeddings and $h_{R,\leq 0}^{(j)}$ the observed-RGB subset. Both visual streams and the action tower receive the encoded task context $C$ through cross-attention. With modality-specific noise levels $\sigma_R,\sigma_D,\sigma_A$, the three feature sequences update synchronously as
\begingroup
\setlength{\jot}{0pt}
\setlength{\abovedisplayskip}{4pt plus 1pt minus 1pt}
\setlength{\belowdisplayskip}{4pt plus 1pt minus 1pt}
\setlength{\belowdisplayshortskip}{4pt plus 1pt minus 1pt}
\begin{equation}
\begin{aligned}
    h_R^{(j+1)} &= \mathcal B_V^{(j)}\!\left(h_R^{(j)}\mid C,\sigma_R\right),\\
    h_D^{(j+1)} &= \mathcal B_V^{(j)}\!\left(h_D^{(j)}\mid h_{R,\leq 0}^{(j)};\,C,\sigma_D\right),\\
    h_A^{(j+1)} &= \mathcal B_A^{(j)}\!\left(h_A^{(j)}\mid h_D^{(j)};\,C,\sigma_D,\sigma_A\right).
\end{aligned}
\label{eq:tower_evolution}
\end{equation}
\endgroup
Here $\mathcal B_V^{(j)}$ is the visual block shared by the RGB and depth streams, and $\mathcal B_A^{(j)}$ is the independently parameterized action block. The visual streams retain modality-specific input, conditioning, and output projections. After the final layer, visual heads project tokens back to latent grids to predict RGB and depth flow velocities, while the action head predicts a velocity vector for each step in normalized action space. These outputs are supervised by the joint flow-matching objective below.

\paragraph{Training Details.}
We initialize the depth VAE from the pretrained Wan VAE and pretrain it on 2,300 hours of depth data (see Appendix~\ref{appn:depth_vae} for the data mixture, losses, and optimization settings); both VAEs and T5 remain frozen during policy training. Policy training follows two stages. First, we adapt the pretrained Wan backbone to depth space by predicting depth sequences conditioned on current depth and language, using depth-prediction loss alone. We then jointly optimize the RGB and depth streams, action tower, and conditioning projections with sparse RGB goal supervision, dense intermediate depth supervision, and action supervision, connecting semantic planning, physical rollout, and executable action generation.

For joint training, let $R_{\leq 0},R_F$ and $D_0,D_{1:F}$ denote the encoded RGB and depth observations and targets, and $A_{1:H}$ the normalized action chunk; visual subscripts identify the source frames rather than the number of temporally compressed latent frames. For targets $(X_R,X_D,X_A)=(R_F,D_{1:F},A_{1:H})$, we independently sample Gaussian noise $\epsilon_m$ and noise levels $\sigma_m\in[0,1]$ across modalities:
\begin{equation}
    \widetilde X_m=(1-\sigma_m)X_m+\sigma_m\epsilon_m,
    \qquad v_m=\epsilon_m-X_m,
    \qquad \epsilon_m\sim\mathcal N(0,I).
\label{eq:corruption}
\end{equation}
Conditioned on $R_{\leq 0},D_0,C$, the model predicts velocities $\widehat v_m$ and minimizes the joint flow-matching objective~\citep{flowmatching}:
\begin{equation}
    \mathcal L=\mathcal L_R+\mathcal L_D+\mathcal L_A,
    \qquad
    \mathcal L_m=\mathbb E\!\left[
       w(\sigma_m)\operatorname{MSE}(\widehat v_m,v_m)
    \right].
\label{eq:joint_loss}
\end{equation}
Here $w$ weights noise levels, and the expectation is over training examples, noise, and noise levels. MSE is averaged over valid prediction targets, excluding observations and padding. The noise schedule and weighting function are specified in Appendix~\ref{appn:implementation}; benchmark-specific budgets, data mixtures, and optimizer settings are collected in \cref{tab:prismwam_training_recipes}.

\paragraph{Inference.}
At inference (\autoref{fig:method_architecture}, lower right), the RGB stream encodes $r_{\leq 0}$ once and caches its features for semantic guidance, without generating future RGB frames. Future depth latents and the normalized action chunk are initialized from Gaussian noise and jointly denoised, with the observed RGB, current depth, and task context held fixed. The resulting depth latents can be decoded into a depth sequence for visualization, but action generation uses intermediate rollout features and does not require decoding depth maps. We denormalize the predicted action chunk, execute a prefix, and replan from the next observation.

\section{Experiment}

Without embodied action pretraining, we evaluate PrismWAM across five simulation benchmarks spanning both in-distribution manipulation and out-of-distribution transfer (\S\ref{sec:main_results}). We further examine how our role-decoupled representations support control through robustness analyses, controlled ablations, and discussion (\S\ref{sec:predictive_state_analysis}), and validate the approach in real-world manipulation (\S\ref{sec:realworld}). Training settings and detailed benchmark results are provided in Appendices~\ref{appn:implementation} and~\ref{appn:benchmark_details}, respectively.

\begin{figure}[htbp]
    \centering
    \includegraphics[width=\textwidth]{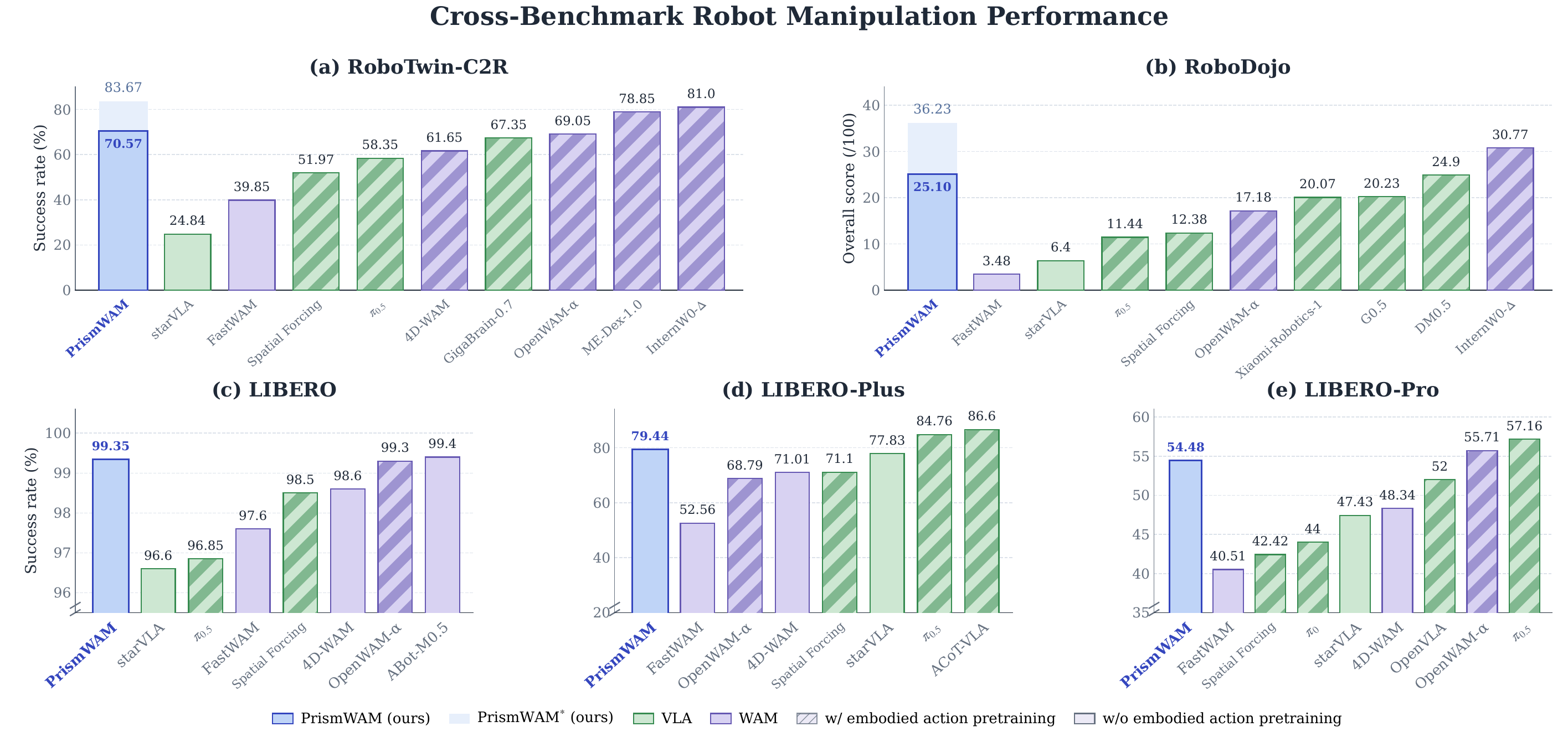}
    \vspace{-2mm}
    \caption{\textbf{Cross-benchmark performance of PrismWAM and PrismWAM$^{*}$.} PrismWAM$^{*}$ denotes our exploratory scaled-up variant. Across five manipulation benchmarks, our methods outperform the cross-benchmark baselines starVLA and FastWAM and remain competitive with leading benchmark-specific methods and models using large-scale embodied action pretraining.}
    \label{fig:benchmark_overview}
    \vspace{-2mm}
\end{figure}

\subsection{Experiment Setup}

\paragraph{Benchmarks.}
We evaluate in-distribution manipulation and out-of-distribution transfer across five simulation benchmarks: RoboTwin clean-to-random~\citep{RoboTwin}, LIBERO~\citep{libero}, LIBERO-Plus~\citep{libero-plus}, LIBERO-Pro~\citep{libero-pro}, and RoboDojo~\citep{robodojo2026}. LIBERO assesses in-distribution manipulation under its standard protocol. RoboTwin clean-to-random, LIBERO-Plus, and LIBERO-Pro probe transfer across diverse visual, spatial, semantic, task, and environmental shifts, while RoboDojo evaluates manipulation capabilities using both task scores and success rates.

\paragraph{Baselines.}
We compare PrismWAM with representative cross-benchmark VLAs and WAMs, including methods trained without embodied action pretraining (starVLA~\citep{starvla}, FastWAM~\citep{fastwam}), methods that use embodied action pretraining ($\pi_{0.5}$~\citep{pi05}, OpenWAM-$\alpha$~\citep{openwam}), and methods that incorporate spatial information during pretraining or fine-tuning (X-WAM~\citep{xwam}, Spatial-Forcing~\citep{spatial-forcing}, 4D-WAM~\citep{4d-wam}). Hatched bars in \cref{fig:benchmark_overview} denote pretrained models and methods fine-tuned from them. We additionally include top-performing benchmark-specific methods~\citep{gigabrain-0.7,abot-m0.5,acot-vla,openvla,pi_0,g05,xiaomi-robotics-1,dm05}. For RoboTwin-C2R, \cref{fig:benchmark_overview} also includes ME-Dex-1.0~\citep{me-dex-1.0}, a leading entry on the official leaderboard as of October 19, 2026~\citep{robotwin-leaderboard2026}. ME-Dex-1.0 uses tactile supervision during training and zero tactile input at evaluation. We also include InternW0-$\Delta$~\citep{internw0-delta}, with publicly available fine-tuning and inference code and benchmark checkpoints. We use its reported clean/random average for RoboTwin-C2R and the October 17, 2026 leaderboard score for RoboDojo~\citep{robodojo-leaderboard2026}.

\subsection{Main Results}
\label{sec:main_results}

\definecolor{benchmint}{HTML}{E8F2EE}
\definecolor{benchminttext}{HTML}{17675F}
\definecolor{benchteal}{HTML}{E4EFF0}
\definecolor{benchtealtext}{HTML}{0D6670}
\definecolor{benchblue}{HTML}{E7EEF6}
\definecolor{benchbluetext}{HTML}{225B7A}

\newcommand{\benchmarkgroup}[2]{%
    \rowcolor{black!7}\multicolumn{#1}{l}{\textit{#2}}\\%
}
\newcommand{\benchsubhead}[3]{%
    \cellcolor{#1}\textcolor{#2}{\scriptsize\bfseries\itshape\strut #3}%
}
\newcommand{\benchmissing}{\textcolor{black!38}{--}}
\newcommand{\secondbest}[1]{\underline{#1}}

\begin{table*}[t]
    \centering
    \setlength{\abovecaptionskip}{5pt}
    \caption{\textbf{RoboTwin-C2R and RoboDojo performance.} RoboTwin-C2R: success (\%); RoboDojo: Score (/100) / SR (\%). PrismWAM$^{*}$ denotes our exploratory scaled-up variant. RoboDojo results for both PrismWAM variants are means over seeds 0--2. \textbf{Bold}: best; \underline{underline}: second best.}
    \label{tab:robotwin_robodojo_comparison}
    \label{tab:standard_benchmark_comparison}
    \label{tab:robotwin_comparison}
    \label{tab:robodojo_comparison}
    \begingroup
    \renewcommand{\arraystretch}{1.0}
    \setlength{\tabcolsep}{3.2pt}
    \setlength{\aboverulesep}{0.3ex}
    \setlength{\belowrulesep}{0.4ex}
    \footnotesize
    \resizebox{\textwidth}{!}{%
    \begin{tabular}{l r r >{\columncolor{black!3}}r r r r r r >{\columncolor{black!3}}r}
        \toprule
        & \multicolumn{3}{>{\columncolor{benchmint}}c}{\textcolor{benchminttext}{\itshape\bfseries RoboTwin-C2R}}
        & \multicolumn{6}{>{\columncolor{benchblue}}c}{\textcolor{benchbluetext}{\itshape\bfseries RoboDojo}} \\
        \cmidrule(lr){2-4}\cmidrule(lr){5-10}
        \textbf{Method}
        & \benchsubhead{benchmint}{benchminttext}{Clean}
        & \benchsubhead{benchmint}{benchminttext}{Rand.}
        & \benchsubhead{benchmint}{benchminttext}{Avg.}
        & \benchsubhead{benchblue}{benchbluetext}{Gen.}
        & \benchsubhead{benchblue}{benchbluetext}{Precision}
        & \benchsubhead{benchblue}{benchbluetext}{Long-H.}
        & \benchsubhead{benchblue}{benchbluetext}{Memory}
        & \benchsubhead{benchblue}{benchbluetext}{Open}
        & \benchsubhead{benchblue}{benchbluetext}{Overall} \\
        \midrule

        \benchmarkgroup{10}{w/ Embodied Action Pretraining}
        $\pi_{0.5}$ & 70.70 & 46.00 & 58.35 & 13.38/8.17 & 12.40/5.50 & 23.54/14.67 & 5.89/4.67 & \secondbest{1.98}/\secondbest{1.67} & 11.44/6.93 \\
        OpenWAM-$\alpha$ & \secondbest{89.40} & 48.70 & 69.05 & 20.71/14.83 & 18.45/9.25 & 34.93/\secondbest{25.33} & 10.41/9.11 & 1.41/1.08 & 17.18/11.92 \\
        \benchmarkgroup{10}{w/o Embodied Action Pretraining}
        starVLA & 46.52 & 3.16 & 24.84 & 3.94/2.34 & 9.90/4.33 & 14.15/6.50 & 3.34/2.44 & 0.67/0.58 & 6.40/3.24 \\
        FastWAM & 77.80 & 1.90 & 39.85 & 2.34/1.11 & 1.96/0.00 & 9.14/5.17 & 3.55/3.44 & 0.42/0.42 & 3.48/2.03 \\
        \textbf{PrismWAM (ours)} & 87.98 & \secondbest{53.16} & \secondbest{70.57} & \secondbest{27.01}/\secondbest{19.56} & \secondbest{28.70}/\secondbest{19.50} & \secondbest{37.39}/22.08 & \secondbest{30.46}/\secondbest{30.11} & 1.92/1.50 & \secondbest{25.10}/\secondbest{18.55} \\
        \textbf{PrismWAM$^{*}$ (ours)} & \textbf{91.84} & \textbf{75.50} & \textbf{83.67} & \textbf{36.71}/\textbf{29.33} & \textbf{39.82}/\textbf{31.08} & \textbf{48.67}/\textbf{31.00} & \textbf{50.96}/\textbf{50.67} & \textbf{4.98}/\textbf{4.42} & \textbf{36.23}/\textbf{29.30} \\
        \benchmarkgroup{10}{w/ Spatial Information}
        X-WAM & 70.00 & 25.80 & 47.90 & 7.39/3.33 & 6.72/1.83 & 17.47/9.08 & 6.32/4.67 & 0.57/0.25 & 7.69/3.83 \\
        Spatial-Forcing & 77.20 & 26.74 & 51.97 & 14.12/9.34 & 17.32/10.58 & 23.26/14.58 & 5.43/4.11 & 1.78/1.58 & 12.38/8.04 \\
        4D-WAM & 81.50 & 41.80 & 61.65 & 13.35/8.83 & 18.08/10.00 & 23.30/14.00 & 7.38/6.00 & 1.44/1.00 & 12.71/7.97 \\
        \bottomrule
    \end{tabular}%
    }
    \endgroup

    \vspace{0.2em}
    \caption{\textbf{LIBERO performance and transfer.} Success rates (\%).}
    \label{tab:libero_family_comparison}
    \label{tab:generalization_benchmark_comparison}
    \label{tab:libero_comparison}
    \label{tab:libero_plus_comparison}
    \label{tab:libero_pro_comparison}
    \begingroup
    \renewcommand{\arraystretch}{1.0}
    \setlength{\tabcolsep}{3.1pt}
    \setlength{\aboverulesep}{0.3ex}
    \setlength{\belowrulesep}{0.4ex}
    \scriptsize
    \resizebox{\textwidth}{!}{%
    \begin{tabular}{l r r r r >{\columncolor{black!3}}r r r r r r r r >{\columncolor{black!3}}r r r r r r >{\columncolor{black!3}}r}
        \toprule
        & \multicolumn{5}{>{\columncolor{benchmint}}c}{\textcolor{benchminttext}{\itshape\bfseries LIBERO}}
        & \multicolumn{8}{>{\columncolor{benchteal}}c}{\textcolor{benchtealtext}{\itshape\bfseries LIBERO-Plus}}
        & \multicolumn{6}{>{\columncolor{benchblue}}c}{\textcolor{benchbluetext}{\itshape\bfseries LIBERO-Pro}} \\
        \cmidrule(lr){2-6}\cmidrule(lr){7-14}\cmidrule(lr){15-20}
        \textbf{Method}
        & \benchsubhead{benchmint}{benchminttext}{Spatial}
        & \benchsubhead{benchmint}{benchminttext}{Object}
        & \benchsubhead{benchmint}{benchminttext}{Goal}
        & \benchsubhead{benchmint}{benchminttext}{Long}
        & \benchsubhead{benchmint}{benchminttext}{Overall}
        & \benchsubhead{benchteal}{benchtealtext}{Camera}
        & \benchsubhead{benchteal}{benchtealtext}{Robot}
        & \benchsubhead{benchteal}{benchtealtext}{Lang.}
        & \benchsubhead{benchteal}{benchtealtext}{Light}
        & \benchsubhead{benchteal}{benchtealtext}{Bg.}
        & \benchsubhead{benchteal}{benchtealtext}{Noise}
        & \benchsubhead{benchteal}{benchtealtext}{Layout}
        & \benchsubhead{benchteal}{benchtealtext}{Overall}
        & \benchsubhead{benchblue}{benchbluetext}{Object}
        & \benchsubhead{benchblue}{benchbluetext}{Pos.}
        & \benchsubhead{benchblue}{benchbluetext}{Sem.}
        & \benchsubhead{benchblue}{benchbluetext}{Task}
        & \benchsubhead{benchblue}{benchbluetext}{Env.}
        & \benchsubhead{benchblue}{benchbluetext}{Overall} \\
        \midrule
        \benchmarkgroup{20}{w/ Embodied Action Pretraining}
        $\pi_{0.5}$ & 98.8 & 98.2 & 98.0 & 92.4 & 96.85 & \secondbest{71.23} & \textbf{76.06} & 85.82 & \secondbest{97.72} & \secondbest{96.19} & \textbf{87.32} & \textbf{86.23} & \textbf{84.76} & \secondbest{85.40} & \textbf{28.15} & 95.50 & \textbf{25.30} & \secondbest{51.45} & \textbf{57.16} \\
        OpenWAM-$\alpha$ & \textbf{99.6} & \secondbest{99.6} & \textbf{99.8} & \secondbest{98.2} & \secondbest{99.3} & 33.21 & \secondbest{75.87} & 87.57 & 97.02 & 85.59 & 39.85 & 77.38 & 68.79 & 84.45 & \secondbest{26.10} & 95.95 & \secondbest{25.25} & 46.80 & \secondbest{55.71} \\
        \benchmarkgroup{20}{w/o Embodied Action Pretraining}
        starVLA & 97.8 & 98.6 & 96.2 & 93.8 & 96.6 & 64.04 & 57.16 & 85.75 & 93.78 & 96.00 & \secondbest{80.95} & 77.25 & 77.83 & 76.65 & 2.80 & 94.85 & 17.80 & 45.05 & 47.43 \\
        FastWAM & 98.2 & \textbf{100.0} & 97.0 & 95.2 & 97.6 & 16.01 & 46.26 & 72.93 & 81.96 & 55.48 & 39.41 & 66.49 & 52.56 & 73.00 & 0.05 & 82.90 & 14.90 & 31.70 & 40.51 \\
        \textbf{PrismWAM (ours)} & \secondbest{99.4} & \textbf{100.0} & 98.2 & \textbf{99.8} & \textbf{99.35} & 60.60 & 67.61 & \textbf{92.13} & \textbf{99.65} & \textbf{99.54} & 67.46 & \secondbest{81.70} & \secondbest{79.44} & \textbf{88.65} & 6.55 & \textbf{98.55} & 23.45 & \textbf{55.20} & 54.48 \\
        \benchmarkgroup{20}{w/ Spatial Information}
        X-WAM & 96.8 & 98.6 & 95.4 & 95.6 & 96.60 & \textbf{72.92} & 46.32 & 88.48 & 93.61 & 88.29 & 62.90 & 80.79 & 74.80 & 83.95 & 21.20 & \secondbest{97.25} & 17.70 & 45.55 & 53.13 \\
        Spatial-Forcing & \secondbest{99.4} & \secondbest{99.6} & \secondbest{98.8} & 96.0 & 98.5 & 56.85 & 37.94 & 84.39 & 95.36 & 91.82 & 65.90 & 79.02 & 71.10 & 73.95 & 2.05 & 94.45 & 9.80 & 31.85 & 42.42 \\
        4D-WAM & 98.8 & 99.5 & 98.6 & 97.4 & 98.6 & 45.15 & 64.26 & \secondbest{90.63} & 94.29 & 57.71 & 69.08 & 79.21 & 71.01 & 80.75 & 3.90 & 95.85 & 21.10 & 40.10 & 48.34 \\
        \bottomrule
    \end{tabular}%
    }
    \endgroup
\end{table*}

Across five benchmarks, PrismWAM \emph{substantially outperforms the cross-benchmark baselines starVLA and FastWAM without embodied action pretraining}. On RoboTwin-C2R, PrismWAM achieves 53.16\% success in randomized scenes, compared with starVLA at 3.16\% (+50.00 pp) and FastWAM at 1.90\% (+51.26 pp). These results demonstrate the strong \emph{zero-shot generalization} capability of PrismWAM without embodied action pretraining. PrismWAM further outperforms the available geometry-aware baselines on LIBERO, reaching 99.35\% versus 98.60\% for 4D-WAM (\cref{tab:libero_family_comparison}). These results suggest that \emph{the key to effectively leveraging physical information is not only to incorporate it, but also to decouple it from semantic representations}.
Moreover, PrismWAM \emph{surpasses both pretrained and non-pretrained WAM baselines} in most settings, achieving 70.57\% clean/random average success on RoboTwin-C2R compared with 69.05\% for pretrained OpenWAM-$\alpha$. On RoboDojo, PrismWAM$^{*}$ achieves an overall score of 36.23 (\cref{fig:benchmark_overview}).
Beyond WAM comparisons, PrismWAM remains competitive with large-scale pretrained policies (\cref{fig:robodojo_pretraining_budget}).
Results on RoboDojo also suggest our \emph{data-efficiency}, achieving strong control without additional embodied backbone pretraining.

\subsection{Analysis and Discussion}
\label{sec:predictive_state_analysis}

\paragraph{RQ1: Does physical-state prediction offer more than appearance robustness?}
Depth encodes distance rather than texture or illumination, making it naturally insensitive to changes in color and lighting. This raises a key question: does physical-state prediction offer more than appearance robustness? Results across multiple benchmarks point to broader benefits for generalization and precise control. PrismWAM achieves 81.70\% under layout perturbations on LIBERO-Plus, compared with 77.38\% for OpenWAM-$\alpha$ and 66.49\% for FastWAM; on LIBERO-Pro, it achieves 88.65\% when object attributes such as size, shape, and color change, exceeding OpenWAM-$\alpha$ at 84.45\% and FastWAM at 73.00\% (\cref{tab:libero_family_comparison}). On RoboDojo's challenging Precision tasks, PrismWAM substantially outperforms pretrained OpenWAM-$\alpha$, achieving 19.50\% success versus 9.25\% (\cref{tab:robotwin_robodojo_comparison}). These results suggest that physical-state prediction goes beyond appearance robustness: \emph{it supports broader generalization, as illustrated by robustness to changes in scene layout and object shape, and exposes physical structure for precise control}.

\begin{figure}[t]
    \centering
    \includegraphics[width=\linewidth]{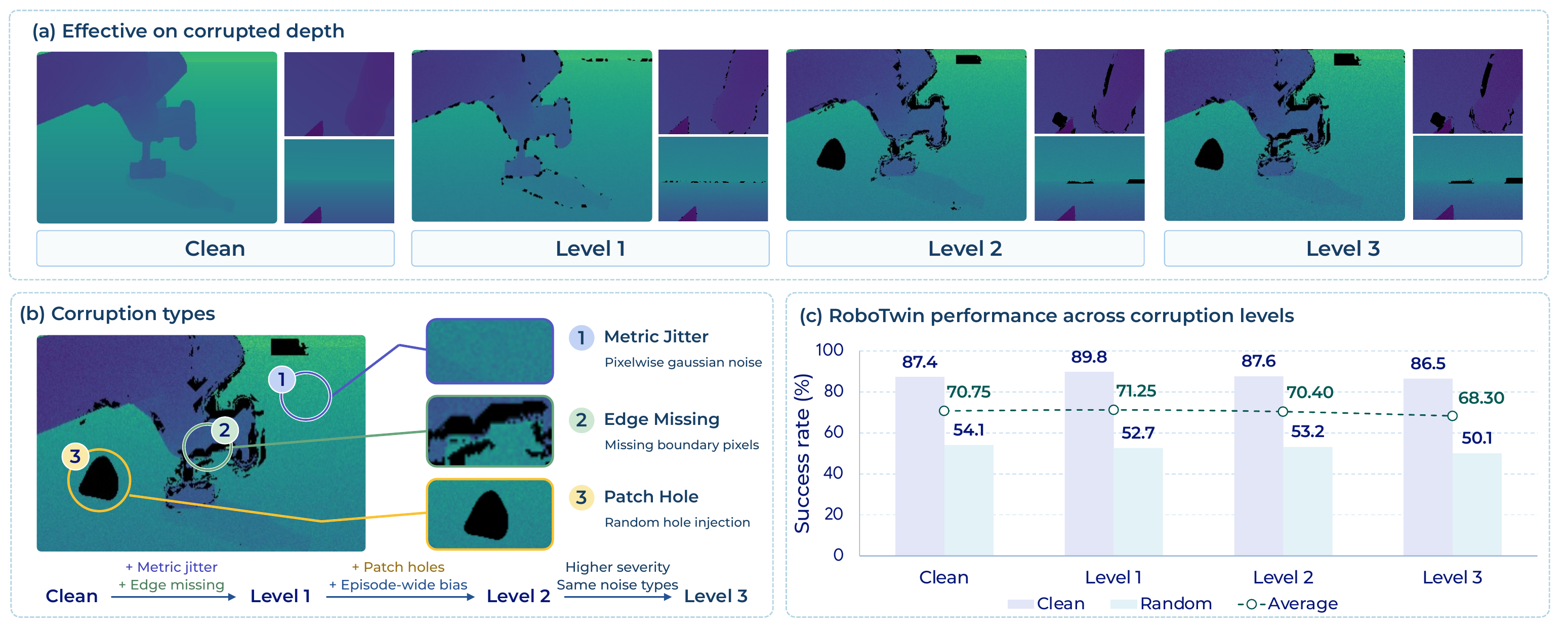}
    \caption{\textbf{Robustness to progressive depth corruption.} \textbf{Top:} Depth observations across corruption levels. \textbf{Bottom left:} Corruption types and their level-wise composition. \textbf{Bottom right:} RoboTwin performance under progressively increasing corruption of the original depth observations.}
    \label{fig:depth_corruption_robustness}
\end{figure}

\paragraph{RQ2: Can PrismWAM remain reliable with real-world-like corruption?}
Real-world depth sensors may produce measurement noise and missing values, motivating evaluation with real-world-like depth corruption. The corruption profiles, which are informed by the manufacturers' specifications for Orbbec Gemini 335 and RealSense D435, are controlled stress tests rather than calibrated models of either sensor; full settings are provided in \cref{appn:depth_corruption}. As shown in \cref{fig:depth_corruption_robustness}, we train and evaluate PrismWAM on RoboTwin-C2R across three corruption levels constructed from \emph{Metric Jitter}, \emph{Edge Missing}, and \emph{Patch Hole}. On randomized scenes, all three corruption levels reduce success relative to uncorrupted depth. In the clean setting, light corruption yields a small improvement before performance declines at higher severity. Nevertheless, even at the strongest corruption level, the average performance drops by only 2.45 pp. These results demonstrate that PrismWAM remains robust to diverse forms of depth corruption.

\paragraph{RQ3: Do PrismWAM's design choices work together effectively?}
We examine the contributions of \emph{representation composition}, \emph{attention routing}, and \emph{supervision and training} through ablations on RoboTwin-C2R (\cref{tab:prismwam_ablation_results}). Full settings and results for all six ablations are reported in Appendix~\ref{appn:ablation_details}.
Removing either the RGB or depth stream reduces both clean-scene success and transfer, supporting the complementary roles of semantic context and physical state.
Direct RGB-to-action access improves clean-scene success but weakens transfer, while exposing depth to future RGB also reduces transfer, favoring the proposed chain from observed semantic context through physical rollout to action.
Together, these results support \emph{representation decoupling through complementary streams}, with the full design achieving the best clean--random average.

\begin{table}[tbp]
    \centering
    \begingroup
    \captionsetup{font=footnotesize,labelfont=normalfont,textfont=normalfont,skip=3pt}
    \newsavebox{\prismAblationPanel}
    \sbox{\prismAblationPanel}{%
        \fontsize{8.5}{10}\selectfont%
        \setlength{\tabcolsep}{3pt}%
        \renewcommand{\arraystretch}{1.24}%
        \setlength{\aboverulesep}{0.3ex}%
        \setlength{\belowrulesep}{0.4ex}%
        \begin{tabular}{@{}p{\dimexpr0.53\linewidth-96pt\relax} >{\centering\arraybackslash}p{26pt} >{\centering\arraybackslash}p{26pt} >{\columncolor{black!3}[\tabcolsep][0pt]\centering\arraybackslash}p{26pt}@{}}
            \toprule
            \textbf{Variant}
            & \benchsubhead{benchmint}{benchminttext}{Clean}
            & \benchsubhead{benchmint}{benchminttext}{Random}
            & \multicolumn{1}{>{\columncolor{benchmint}[\tabcolsep][0pt]}c@{}}{\textcolor{benchminttext}{\scriptsize\bfseries\itshape\strut Avg.}} \\
            \midrule
            \textbf{PrismWAM} & 87.40 & 54.10 & 70.75 \\
            \benchmarkgroup{4}{Representation composition}
            w/o RGB stream & 79.10 & 40.10 & 59.60 \\
            w/o depth stream & 83.00 & 2.40 & 42.70 \\
            \benchmarkgroup{4}{Attention routing}
            Observed RGB $\rightarrow$ Action & 89.10 & 40.30 & 64.70 \\
            Future RGB $\rightarrow$ Depth & 87.00 & 50.00 & 68.50 \\
            \benchmarkgroup{4}{Supervision \& training}
            w/o future-RGB supervision & 81.60 & 51.40 & 66.50 \\
            w/o depth-only stage & 89.30 & 24.70 & 57.00 \\
            \bottomrule
        \end{tabular}%
    }
    \begin{minipage}[t]{0.53\linewidth}
        \vspace{0pt}
        \caption{Component ablations on 50 RoboTwin tasks, with 20 episodes per task and condition.}
        \label{tab:prismwam_ablation_results}
    \end{minipage}\hfill
    \begin{minipage}[t]{0.45\linewidth}
        \vspace{0pt}
        \captionof{figure}{RoboDojo score and success rate.}
        \label{fig:robodojo_pretraining_budget}
        \label{tab:robodojo_pretraining_budget}
    \end{minipage}\par\nointerlineskip
    \begin{minipage}[t]{0.53\linewidth}
        \vspace{0pt}
        \centering
        \usebox{\prismAblationPanel}
    \end{minipage}\hfill
    \begin{minipage}[t]{0.45\linewidth}
        \vspace{0pt}
        \centering
        \includegraphics[height=\dimexpr\ht\prismAblationPanel+\dp\prismAblationPanel\relax,trim=0 2.5bp 0 1.75bp,clip]{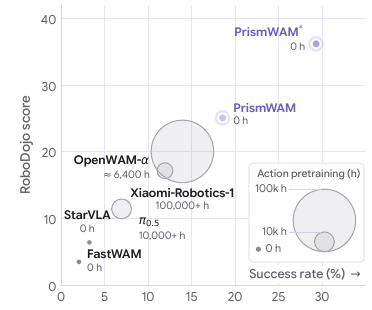}
    \end{minipage}
    \endgroup
\end{table}

\subsection{Real-world Validation}\label{sec:realworld}

\begin{figure}[t]
    \centering
    \includegraphics[width=\linewidth]{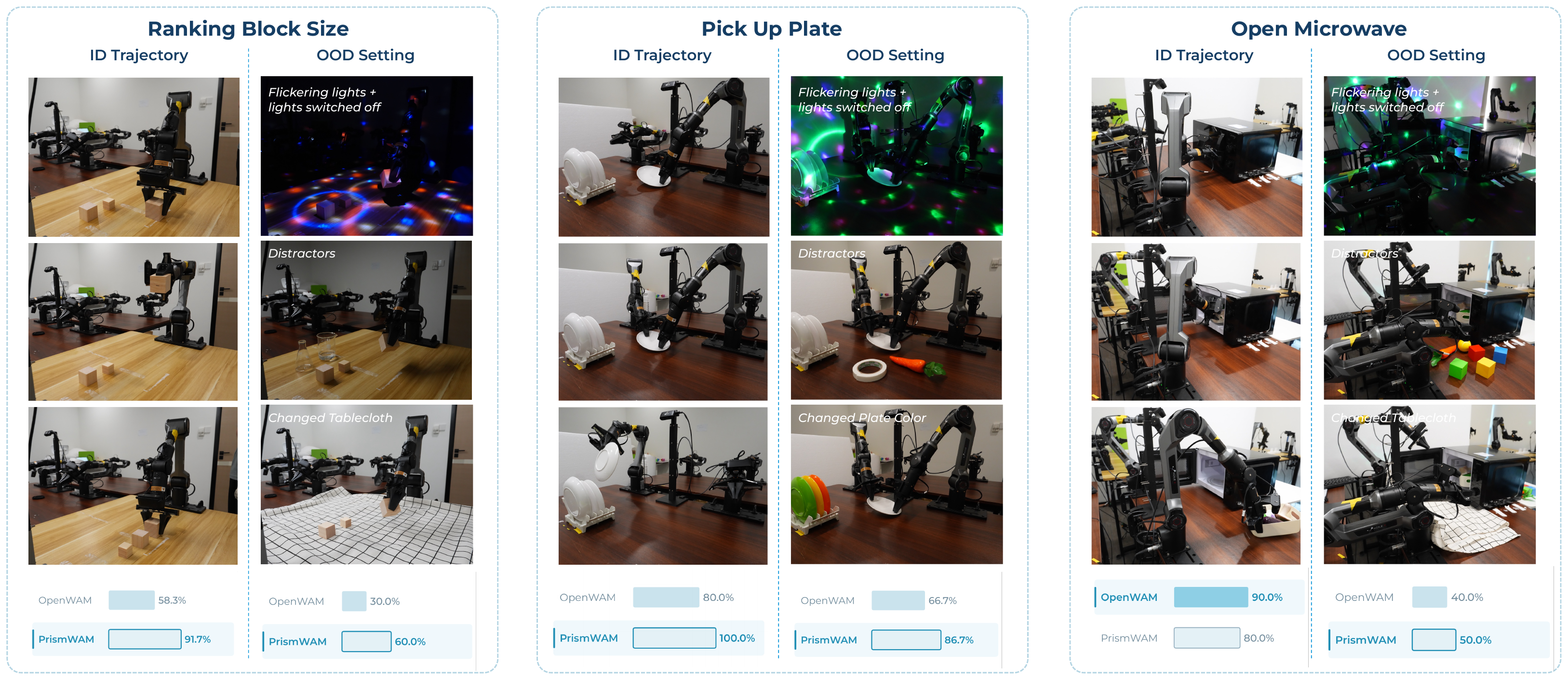}
    \caption{\textbf{Real-world evaluation.} ID trajectories (left), representative OOD scenes (right), and OpenWAM versus PrismWAM success rates (bottom) across three tasks.}
    \label{fig:realworld_execution}
\end{figure}

\paragraph{Tasks and evaluation conditions.}
We compare PrismWAM with OpenWAM~\citep{openwam} on three real-robot tasks: \emph{Pick Up Plate}, \emph{Ranking Block Size}, and \emph{Open Microwave}. These tasks cover bimanual plate handling, single-arm size-dependent ordering, and multistep bimanual retrieval, respectively. In-distribution (ID) trials use the original setups and task-specific initial arrangements. Out-of-distribution (OOD) trials vary illumination, distractors, and object appearance or placement while preserving each task's goal. \Cref{fig:realworld_execution,tab:realworld_results} summarize the settings and results. Task instructions, training recipes, are detailed in Appendix~\ref{appn:realworld_details}.

\begin{table}[t]
\small
    \centering
    \caption{\textbf{Real-world success under ID and OOD conditions.} Entries are successes/trials (\%). Both models are evaluated on the same three tasks and trial counts. Pooled rates divide total successes by total trials across tasks. Bold marks the higher success rate within each task and condition.}
    \label{tab:realworld_results}
    \begingroup
    \footnotesize
    \setlength{\tabcolsep}{3pt}
    \renewcommand{\arraystretch}{1.1}
    \begin{tabular*}{\linewidth}{@{\extracolsep{\fill}}lcccc@{}}
        \toprule
        & \multicolumn{2}{c}{\textbf{OpenWAM}} & \multicolumn{2}{c}{\textbf{PrismWAM}} \\
        \cmidrule(lr){2-3} \cmidrule(lr){4-5}
        \textbf{Task} & ID & OOD & ID & OOD \\
        \midrule
        Pick Up Plate & 4/5 (80.0) & 10/15 (66.7) & \textbf{5/5 (100.0)} & \textbf{13/15 (86.7)} \\
        Ranking Block Size & 7/12 (58.3) & 3/10 (30.0) & \textbf{11/12 (91.7)} & \textbf{6/10 (60.0)} \\
        Open Microwave & \textbf{9/10 (90.0)} & 4/10 (40.0) & 8/10 (80.0) & \textbf{5/10 (50.0)} \\
        \midrule
        Pooled (3 tasks) & 20/27 (74.1) & 17/35 (48.6) & \textbf{24/27 (88.9)} & \textbf{24/35 (68.6)} \\
        \bottomrule
    \end{tabular*}
    \endgroup
    \vspace{-3mm}
\end{table}

PrismWAM achieves pooled success of 88.9\% in ID and 68.6\% in OOD, compared with 74.1\% and 48.6\% for OpenWAM, improving success by 14.8 and 20.0 percentage points, respectively. OOD gains hold across all three tasks: 86.7\% versus 66.7\% on Pick Up Plate, 60.0\% versus 30.0\% on Ranking Block Size, and 50.0\% versus 40.0\% on Open Microwave. In ID, PrismWAM improves plate handling and size ordering, while OpenWAM performs better on microwave retrieval (90.0\% versus 80.0\%). These results extend the scene-change robustness observed in simulation to physical manipulation, supporting the practical value of combining RGB semantic guidance with depth-based physical rollouts.

\section{Conclusion}
We introduce \textbf{PrismWAM}, a World Action Model built on the principle of decoupling semantic context from physical state. Rather than treating video latents as a monolithic world representation, PrismWAM decomposes world-action modeling into $\mathrm{Semantic~Planning}\rightarrow\mathrm{Physical~Rollout}\rightarrow\mathrm{Executable~Action}$: RGB future-frame prediction captures task-level intent, depth models physical state transitions, and the resulting rollout is mapped directly to continuous actions without RGB goal prediction at inference. Across diverse simulation and real-world manipulation benchmarks, PrismWAM achieves strong performance and generalization with only 2{,}300 hours of depth data and without embodied action pretraining, while matching or surpassing large-scale action-pretrained models on challenging settings. These results suggest that scaling action data is not the only path to stronger embodied foundation models; principled state representations and architectural design offer a more data-efficient alternative. More broadly, PrismWAM paves the way for a new generation of WAMs that shifts the focus from modeling \textit{how the world looks} to \textit{how the world responds to action}. While depth is one practical realization of this principle, it can extend to occupancy, flow, tactile sensing, proprioception, and other representations capturing action-induced state transitions.

\paragraph{Limitations and future work.}
At present, large-scale real-world manipulation data with synchronized depth observations remains relatively limited, which constrains the scale of depth pretraining compared with existing action-pretrained models. We view this primarily as a reflection of the current data landscape: as depth sensing becomes more accessible and embodied datasets incorporate richer physical modalities, such data is likely to become increasingly available. Physical-state pretraining may therefore emerge as a natural scaling direction for World Action Models, with PrismWAM taking an early step toward establishing this new paradigm.

\subsection*{AI use statement}

Generative AI tools were used to assist with manuscript drafting, language polishing, and structural editing, as well as with literature search and code development. 
AI-generated suggestions were used only as assistance and were reviewed, revised, and verified by the authors. All technical claims, experimental results, interpretations, and final written content were validated by the authors, who take full responsibility for the work.

\section*{Acknowledgments}
We thank Chong Kar Hing, Junwei Li, and Zexuan Wen for their assistance with the experiments, and Shihao Wang and Yu-Cheng Chou for their insightful ideas.

\bibliography{iclr2027_conference}
\bibliographystyle{iclr2027_conference}

\appendix

\clearpage
\section{Training and Additional Experiments}
\label{appn:experimental_details}
\begingroup
\setlength{\parskip}{3pt}
\captionsetup{font=small,skip=3pt}

\subsection{PrismWAM Training Recipe}
\label{appn:implementation}
\begingroup
\setlength{\parskip}{3pt}

The main PrismWAM model has approximately 6B parameters and uses Wan2.2-5B~\citep{wan2.2}; PrismWAM$^{*}$ denotes the exploratory 15B variant, which is initialized from WAN-2.1-14B with additional video-only pretraining. Training comprises depth-only adaptation conditioned on current depth and language, followed by joint RGB/depth/action training; both VAEs and T5 remain frozen.

\noindent\begin{minipage}{\linewidth}
\captionsetup{font=small,skip=3pt}
\captionof{table}{PrismWAM training and deployment settings. C2R: clean-to-random transfer; EE: end-effector; $H/K$: action horizon/executed prefix.}
\label{tab:prismwam_training_recipes}
\label{tab:prismwam_shared_defaults}
\centering
\small
\setlength{\tabcolsep}{3pt}
\renewcommand{\arraystretch}{1.02}
\begin{tabular*}{\linewidth}{@{\extracolsep{\fill}}l*{3}{c}@{}}
\toprule
Setting & \shortstack{RoboTwin\\C2R} & LIBERO & RoboDojo \\
\midrule
Training data & Clean & Four suites & ARX X5 sim. \\
Tasks / total demos & 50 / 2,500 & 40 / 2,000 & 35 / 3,500 \\
Demos per task & 50 clean & 50 & 100 \\
\midrule
Optimizer updates: depth / joint & 15k / 15k & 15k / 15k & 15k / 40k \\
Global batch size & 256 & 256 & 1024 \\
\midrule
Views; mosaic $(H\times W)$ & $3;\ 384\times320$ & $2;\ 224\times448$ & $3;\ 384\times320$ \\
Depth frames & 9 & 5 & 9 \\
RGB history frames & 5 & 5 & 9 \\
Future-RGB target & $t+32$ & $t+16$ & $t+32$ \\
Action representation & 14-D absolute joints & 7-D EE deltas & 14-D abs.\ joints \\
Proprioceptive dimension & 14 & 8 & 14 \\
Target action indices & $[t+1,t+32]$ & $[t,t+15]$ & $[t+1,t+32]$ \\
Action / state statistics & Shared mean/std & Separate min--max & Shared mean/std \\
Horizon / replan $(H/K)$ & 32 / 24 & 16 / 16 & 32 / 24 \\
\bottomrule
\end{tabular*}
\end{minipage}

\textbf{Optimization and evaluation.}
AdamW uses $\beta=(0.9,0.95)$, $\epsilon=10^{-8}$, weight decay 0.01, gradient clipping 1.0, and bf16 without gradient accumulation. Each stage resets the optimizer; all trainable modules share 5\% linear warmup and cosine decay from $10^{-4}$ to $10^{-6}$. We use seed 42, uniform task/window sampling, and checkpoints every 1000 updates. Main results use the final checkpoint and no sensor-depth corruption. LIBERO trains on Spatial/Object/Goal/Long; Plus/Pro reuse its checkpoint. Inference uses ten shift-5 Euler steps with cached observed RGB.

\textbf{Inputs.}
Three-view mosaics place a $256\times320$ head image above two $128\times160$ wrist images; LIBERO places $224\times224$ external/wrist images left/right. RGB/depth layouts are shared across training and evaluation. RoboTwin/LIBERO encode RGB history as one causal clip (two latent frames) and the future target separately (one latent frame); RoboDojo uses history-noise probability 0.25. Depth clips include the current frame; actions use consecutive steps. Out-of-range frames repeat boundary observations, with padded targets masked.

\textbf{Flow matching.}
Independently per predicted modality, sample $u\sim\mathcal U(0,1)$, set $\sigma=5u/(1+4u)$, and use timestep $1000\sigma$. For the targets in \cref{eq:corruption,eq:joint_loss},
\begin{equation}
    w(\sigma)=\frac{y(\sigma)-y_{\min}}{\overline y-y_{\min}+10^{-10}},
    \qquad y(\sigma)=\exp[-2(\sigma-1/2)^2],
    \label{eq:appendix_flow_weight}
\end{equation}
where $y_{\min}$ and $\overline y$ are the minimum and mean on the 1000-point shifted grid $u\in\{1,0.999,\ldots,0.001\}$. Modality losses have unit coefficients and exclude conditioning frames and padded targets.
\par\endgroup

\clearpage
\subsection{Depth-VAE Training Recipe}
\label{appn:depth_vae}
We train Wan2.2-compatible and Wan2.1-compatible single-channel depth VAEs with 48 and 16 latent channels, respectively. Each is initialized from its corresponding pretrained RGB VAE~\citep{wan2.2}, averaging the RGB input and output boundary weights across channels while retaining matching internal weights. Their spatial compression factors are 16 and 8, respectively. We additionally train a head-view-only depth VAE for real-world setups without wrist-view depth. The depth VAEs remain frozen throughout downstream policy training.

\paragraph{Training data and mixture.}
We use processed AgiBotWorld-Beta depth~\citep{agibotworld} and simulated RoboTwin depth from clean and randomized demonstrations~\citep{RoboTwin}. After additional filtering of AgiBotWorld-Beta, the combined corpus contains approximately 2,300 hours of depth sequences. We sample AgiBotWorld-Beta and RoboTwin with probabilities 0.75 and 0.25, respectively, then sample a valid temporal clip with replacement. Training augmentation comprises random temporal cropping, in-plane rotation within $\pm5^\circ$, and horizontal flipping with probability 0.5, applied consistently across frames and views.

\Cref{tab:depth_vae_recipes} lists the shared training settings. The training objective is
\begin{equation}
    \mathcal L_{\mathrm{VAE}}=
    \lVert\hat d-d\rVert_1+
    \lVert\hat d-d\rVert_2^2+
    10^{-6}\,D_{\mathrm{KL}}\!\left(q(z\mid d)\,\|\,\mathcal N(0,I)\right),
    \label{eq:depth_vae_objective}
\end{equation}
with mean-reduced reconstruction terms, stochastic posterior sampling, and no additional valid-depth mask. Per-channel latent means and standard deviations are computed from encoder posterior means and stored with each checkpoint for downstream normalization.

\noindent\begin{minipage}{\linewidth}
    \centering
    \captionof{table}{Shared depth-VAE training settings.}
    \label{tab:depth_vae_recipes}
    \small
    \setlength{\tabcolsep}{8pt}
    \renewcommand{\arraystretch}{1.08}
    \begin{tabular}{ll}
        \toprule
        Setting & Value \\
        \midrule
        Clip length / frame stride & 9 frames / 4 \\
        Temporal compression factor & 4 \\
        Depth clipping / normalization & 0--1500\,mm / $[-1,1]$ \\
        Global batch / gradient accumulation & 256 / 1 \\
        Optimizer & AdamW, $\beta=(0.9,0.95)$ \\
        Weight decay / gradient clipping & 0.01 / 1.0 \\
        Precision & bf16 \\
        Warmup updates & 5000 \\
        Learning-rate schedule & Linear warmup, then cosine decay \\
        Selected checkpoint & 50k updates \\
        \bottomrule
    \end{tabular}
\end{minipage}

\subsection{Component Ablations}
\label{appn:ablation_details}
We ablate PrismWAM initialized from Wan2.2-5B, using clean-only RoboTwin demonstrations, global batch 256, and 30k updates (15k depth adaptation plus 15k joint training; 30k joint updates without the depth-only stage). Inference uses ten denoising steps and replans every 24 actions; the future-RGB-to-depth variant jointly denoises RGB, depth, and actions. We evaluate final checkpoints on 50 tasks with 20 episodes per task and Clean/Random condition (1000 each), compared with 100 episodes per task and condition in the main PrismWAM results. Evaluation uses unseen instructions, policy seed 0, instruction seed 42, and environment seeds starting at 100000.

\noindent\begin{minipage}{\linewidth}
\captionof{table}{RoboTwin success (\%); $\Delta$: Mean change from the full model (pp).}
\label{tab:prismwam_ablation_results_appendix}\label{tab:ablation_full}
\centering\small
\setlength{\tabcolsep}{3.2pt}
\renewcommand{\arraystretch}{1.0}
\setlength{\aboverulesep}{0.3ex}
\setlength{\belowrulesep}{0.4ex}
\begin{tabular}{@{}p{\dimexpr0.8\linewidth-166.4pt\relax} >{\raggedleft\arraybackslash}p{35.2pt} >{\raggedleft\arraybackslash}p{35.2pt} >{\columncolor{black!3}\raggedleft\arraybackslash}p{35.2pt} >{\raggedleft\arraybackslash}p{35.2pt}@{}}
\toprule
\textbf{Variant}
& \benchsubhead{benchmint}{benchminttext}{Clean}
& \benchsubhead{benchmint}{benchminttext}{Random}
& \benchsubhead{benchmint}{benchminttext}{Mean}
& \multicolumn{1}{>{\columncolor{benchmint}[\tabcolsep][0pt]}r@{}}{\textcolor{benchminttext}{\scriptsize\bfseries\itshape\strut $\Delta$}} \\
\midrule
\textbf{PrismWAM (full)} & 87.40 & 54.10 & \textbf{70.75} & -- \\
\benchmarkgroup{5}{Representation composition}
w/o RGB stream & 79.10 & 40.10 & 59.60 & $-11.15$ \\
w/o depth stream & 83.00 & 2.40 & 42.70 & $-28.05$ \\
\benchmarkgroup{5}{Attention routing}
Observed RGB $\rightarrow$ Action & 89.10 & 40.30 & 64.70 & $-6.05$ \\
Future RGB $\rightarrow$ Depth & 87.00 & 50.00 & 68.50 & $-2.25$ \\
\benchmarkgroup{5}{Supervision \& training}
w/o future-RGB supervision & 81.60 & 51.40 & 66.50 & $-4.25$ \\
w/o depth-only stage & 89.30 & 24.70 & 57.00 & $-13.75$ \\
\bottomrule
\end{tabular}
\end{minipage}

\clearpage
\subsection{Depth Corruption}
\label{appn:depth_corruption}
\textbf{Perturbation Setting.}
We corrupt metric depth at three severity levels (\cref{fig:depth_corruption_robustness,tab:depth_corruption_settings}). \emph{Metric Jitter} adds Gaussian noise whose standard deviation increases quadratically with depth, ranging from $\sigma_0=12$--$36$\,mm at 0\,m to $\sigma_1=20$--$60$\,mm at 1\,m, with uniform clip-/episode-wise offsets in $[-25,25]$ and $[-50,50]$\,mm at Levels~2 and~3, respectively. \emph{Edge Missing} removes pixels around depth discontinuities exceeding 30\,mm, using dilation radii of 1, 5, or 7 pixels and nominal removal fractions of 24--25\%, modulated by depth at Levels~2--3. \emph{Patch Hole} is enabled at Levels~2--3: each camera receives at most one connected hole targeting 1.5\% of valid pixels per event, with a duration of 4--8 frames and onset offsets of 16--32 frames within each 80-frame cycle.

\noindent\begin{minipage}{\linewidth}
\captionof{table}{Depth corruption parameters and success rates (\%). Edge radius is the native-resolution four-neighbor dilation count; edge fraction is measured within the edge band. Patch area is relative to valid pixels during active events.}
\label{tab:depth_corruption_settings}
\centering\footnotesize
\setlength{\tabcolsep}{3pt}\renewcommand{\arraystretch}{1.08}
\begin{tabular*}{\linewidth}{@{\extracolsep{\fill}}lccccrrrr@{}}
\toprule
Level & \shortstack{$\sigma_0/\sigma_1$\\(mm)} & \shortstack{Shift\\(mm)} & \shortstack{Edge\\radius} & \shortstack{Edge\\fraction} & \shortstack{Patch\\area} & Clean & Random & Mean \\
\midrule
Clean depth & 0/0 & 0 & -- & 0 & 0 & 87.40 & 54.10 & 70.75 \\
1 & 12/20 & 0 & 1 & 25\% & 0 & 89.80 & 52.70 & 71.25 \\
2 & 24/40 & $[-25,25]$ & 5 & 25\% & 1.5\% & 87.60 & 53.20 & 70.40 \\
3 & 36/60 & $[-50,50]$ & 7 & 24\% & 1.5\% & 86.50 & 50.10 & 68.30 \\
\bottomrule
\end{tabular*}
\end{minipage}

\textbf{Training and evaluation.}\label{appn:depth_corruption_training}
Each severity level uses 15k depth-only updates followed by 15k joint updates. Evaluation uses the matching corruption on 50 RoboTwin tasks, with 20 rollouts per task in each Clean/Random condition (1000 per condition; 2000 per level).

\subsection{Real-World Evaluation}
\label{appn:realworld_details}
\textbf{Training.}\label{appn:realworld_training}
Both policies train on the same demonstrations for 30k total updates with a global batch size of 256 and seed 42. PrismWAM is initialized from Wan2.2-5B and uses 15k depth-only adaptation updates followed by 15k joint RGB/depth/action updates; OpenWAM-$\alpha$ uses 30k RGB/action updates. For setups without wrist-view depth, PrismWAM uses the separately pretrained head-view-only depth VAE with $256\times320$ inputs (Appendix~\ref{appn:depth_vae}). The VAEs and T5 remain frozen during policy training. PrismWAM predicts absolute joint commands, and OpenWAM-$\alpha$ predicts absolute end-effector targets.

\textbf{Embodiment and data.}
We use single-arm and dual-arm Piper-X configurations, each equipped with one head-view RGB-D camera (Orbbec Gemini 335). \emph{Ranking Block Size} uses a single arm, while \emph{Pick Up Plate} and \emph{Open Microwave} use both arms. For these three tasks, we collect 100 demonstration episodes for \emph{Open Microwave}, 200 for \emph{Pick Up Plate}, and 180 for \emph{Ranking Block Size}, with 30 episodes for each of the six size permutations. Each task has 50 goal-preserving training paraphrases.

\textbf{Tasks and evaluation.}
Success requires (i) \emph{Pick Up Plate}: pick up the plate with both arms and place it on the rack; (ii) \emph{Ranking Block Size}: place the largest cube farthest from the robot, the medium cube centrally, and the smallest nearest; (iii) \emph{Open Microwave}: open the door, move the inside basket to the tabletop on the microwave's right, and leave the door open. ID uses the original setup, including all six size permutations for \emph{Ranking Block Size}. As illustrated in \cref{fig:realworld_execution}, OOD settings for all three tasks include flickering lights with ambient lights switched off and added distractor objects. Additional settings change plate colors for \emph{Pick Up Plate} and tablecloths for \emph{Ranking Block Size} and \emph{Open Microwave}. Table~\ref{tab:realworld_task_instructions} reports successes/trials; overall rates pool trials.

\noindent\begin{minipage}{\linewidth}
\captionof{table}{Real-world successes/trials.}
\label{tab:realworld_task_instructions}
\centering\small\setlength{\tabcolsep}{7pt}
\begin{tabular*}{\linewidth}{@{\extracolsep{\fill}}lrrrr@{}}
\toprule
 & \multicolumn{2}{c}{OpenWAM-$\alpha$} & \multicolumn{2}{c}{PrismWAM} \\
\cmidrule(lr){2-3}\cmidrule(lr){4-5}
Task & ID & OOD & ID & OOD \\
\midrule
Pick Up Plate & 4/5 & 10/15 & 5/5 & 13/15 \\
Ranking Block Size & 7/12 & 3/10 & 11/12 & 6/10 \\
Open Microwave & 9/10 & 4/10 & 8/10 & 5/10 \\
\bottomrule
\end{tabular*}
\end{minipage}

\par\endgroup

\clearpage
\section{Detailed Results across Five Benchmarks}
\label{appn:benchmark_details}
\begingroup
\setlength{\parskip}{3pt}
\captionsetup{font=small,skip=3pt}

\subsection{LIBERO, LIBERO-Plus, and LIBERO-Pro}
\label{appn:libero_plus_results}
We evaluate PrismWAM on LIBERO~\citep{libero}, LIBERO-Plus~\citep{libero-plus}, and LIBERO-Pro~\citep{libero-pro} with an action horizon of 16 and replanning every 16 actions. LIBERO-Plus and LIBERO-Pro reuse the checkpoint trained on standard LIBERO without additional fine-tuning. For LIBERO-Plus, we apply the RGB sensor-noise settings to depth observations using the same corruption parameters. LIBERO-Pro evaluates five perturbation types with 2000 episodes per type.

\noindent\begin{minipage}{\linewidth}
    \centering
    \captionof{table}{PrismWAM on standard LIBERO and LIBERO-Plus by base suite.}
    \label{tab:libero_episode_counts}
    \small
    \setlength{\tabcolsep}{6pt}
    \renewcommand{\arraystretch}{1.02}
    \begin{tabular}{l r r r r}
        \toprule
        & \multicolumn{2}{c}{\textbf{LIBERO Standard}} & \multicolumn{2}{c}{\textbf{LIBERO-Plus}} \\
        \cmidrule(lr){2-3} \cmidrule(lr){4-5}
        \textbf{Suite} & \textbf{Success / Total} & \textbf{SR (\%)} & \textbf{Success / Total} & \textbf{SR (\%)} \\
        \midrule
        Spatial & 497/500 & 99.40 & 2075/2402 & 86.39 \\
        Object & 500/500 & 100.00 & 2127/2518 & 84.47 \\
        Goal & 491/500 & 98.20 & 1830/2591 & 70.63 \\
        Long & 499/500 & 99.80 & 1936/2519 & 76.86 \\
        \midrule
        \textbf{Overall} & \textbf{1987/2000} & \textbf{99.35} & \textbf{7968/10030} & \textbf{79.44} \\
        \bottomrule
    \end{tabular}
\end{minipage}

\noindent\begin{minipage}{\linewidth}
    \centering
    \captionof{table}{PrismWAM on LIBERO-Plus by perturbation. Overall pools episodes rather than averaging rounded category percentages.}
    \label{tab:libero_plus_category_counts}
    \small
    \setlength{\tabcolsep}{12pt}
    \renewcommand{\arraystretch}{1.02}
    \begin{tabular}{@{}lrr@{}}
        \toprule
        Perturbation & Success / Total & SR (\%) \\
        \midrule
        Camera viewpoints & 969/1599 & 60.60 \\
        Robot initial states & 1048/1550 & 67.61 \\
        Language instructions & 1416/1537 & 92.13 \\
        Light conditions & 1138/1142 & 99.65 \\
        Background textures & 1071/1076 & 99.54 \\
        Sensor noise (matched RGB/depth) & 1080/1601 & 67.46 \\
        Object layout & 1246/1525 & 81.70 \\
        \midrule
        \textbf{Overall} & \textbf{7968/10030} & \textbf{79.44} \\
        \bottomrule
    \end{tabular}
\end{minipage}

\clearpage
\subsection{RoboTwin-C2R: Per-task Results}
\label{appn:taskwise}\label{appn:robotwin_taskwise}
RoboTwin-C2R trains the policy on clean demonstrations only and evaluates 50 tasks in Clean and Random conditions. PrismWAM and PrismWAM$^{*}$ results in \cref{tab:robotwin_per_task} use 100 episodes per task and condition.

\begingroup
\fontsize{8.5}{10}\selectfont\setlength{\tabcolsep}{4pt}\renewcommand{\arraystretch}{1.00}
\begin{longtable}{@{}p{0.42\linewidth}rrrr@{}}
\caption{RoboTwin-C2R per-task SR (\%) after clean-only training.}\label{tab:robotwin_per_task}\label{tab:robotwin_per_task_randomized}\\
\toprule
 & \multicolumn{2}{c}{PrismWAM} & \multicolumn{2}{c}{PrismWAM$^{*}$} \\
\cmidrule(lr){2-3}\cmidrule(lr){4-5}
Task & Clean & Rand. & Clean & Rand. \\
\midrule
\endfirsthead
\multicolumn{5}{l}{\small Table \thetable{} (continued)}\\
\toprule
 & \multicolumn{2}{c}{PrismWAM} & \multicolumn{2}{c}{PrismWAM$^{*}$} \\
\cmidrule(lr){2-3}\cmidrule(lr){4-5}
Task & Clean & Rand. & Clean & Rand. \\
\midrule
\endhead
\midrule\multicolumn{5}{r}{\small Continued on next page}\\\endfoot
\bottomrule\endlastfoot
Adjust Bottle & 98 & 54 & 100 & 97 \\
Beat Block Hammer & 81 & 48 & 98 & 58 \\
Blocks Ranking RGB & 87 & 32 & 93 & 72 \\
Blocks Ranking Size & 56 & 27 & 70 & 55 \\
Click Alarmclock & 100 & 86 & 100 & 98 \\
Click Bell & 100 & 78 & 100 & 93 \\
Dump Bin Bigbin & 84 & 80 & 92 & 82 \\
Grab Roller & 100 & 89 & 100 & 99 \\
Handover Block & 99 & 37 & 98 & 66 \\
Handover Mic & 95 & 24 & 100 & 80 \\
Hanging Mug & 38 & 17 & 59 & 41 \\
Lift Pot & 99 & 67 & 100 & 86 \\
Move Can Pot & 99 & 19 & 100 & 47 \\
Move Pillbottle Pad & 100 & 19 & 100 & 78 \\
Move Playingcard Away & 99 & 65 & 100 & 92 \\
Move Stapler Pad & 33 & 8 & 64 & 33 \\
Open Laptop & 91 & 50 & 94 & 83 \\
Open Microwave & 57 & 19 & 44 & 3 \\
Pick Diverse Bottles & 93 & 78 & 92 & 84 \\
Pick Dual Bottles & 100 & 96 & 100 & 96 \\
Place A2B Left & 83 & 39 & 92 & 74 \\
Place A2B Right & 80 & 41 & 88 & 70 \\
Place Bread Basket & 94 & 77 & 95 & 78 \\
Place Bread Skillet & 90 & 66 & 91 & 74 \\
Place Burger Fries & 98 & 90 & 96 & 100 \\
Place Can Basket & 81 & 31 & 79 & 61 \\
Place Cans Plasticbox & 100 & 95 & 100 & 99 \\
Place Container Plate & 100 & 68 & 100 & 93 \\
Place Dual Shoes & 73 & 43 & 94 & 70 \\
Place Empty Cup & 99 & 70 & 100 & 84 \\
Place Fan & 88 & 42 & 97 & 68 \\
Place Mouse Pad & 89 & 29 & 91 & 43 \\
Place Object Basket & 86 & 44 & 88 & 78 \\
Place Object Scale & 88 & 37 & 97 & 81 \\
Place Object Stand & 97 & 76 & 99 & 90 \\
Place Phone Stand & 88 & 47 & 94 & 75 \\
Place Shoe & 92 & 55 & 99 & 95 \\
Press Stapler & 92 & 50 & 85 & 80 \\
Put Bottles Dustbin & 90 & 82 & 92 & 93 \\
Put Object Cabinet & 89 & 28 & 90 & 52 \\
Rotate QRcode & 83 & 26 & 89 & 49 \\
Scan Object & 94 & 72 & 97 & 81 \\
Shake Bottle & 100 & 79 & 100 & 98 \\
Shake Bottle Horizontally & 100 & 80 & 100 & 97 \\
Stack Blocks Three & 90 & 33 & 95 & 87 \\
Stack Blocks Two & 98 & 60 & 100 & 98 \\
Stack Bowls Three & 84 & 59 & 82 & 79 \\
Stack Bowls Two & 94 & 74 & 96 & 87 \\
Stamp Seal & 80 & 23 & 95 & 33 \\
Turn Switch & 70 & 49 & 67 & 65 \\
\midrule
\textbf{Mean over 50 tasks} & \textbf{87.98} & \textbf{53.16} & \textbf{91.84} & \textbf{75.50} \\
\end{longtable}
\endgroup

\clearpage
\subsection{RoboDojo: Category, Seed, and Per-task Results}
\label{appn:robodojo_taskwise}
The simulation suite~\citep{robodojo2026} has 42 canonical tasks in five categories, represented by 54 standard/randomized entries per seed. Generalization uses 25 episodes/entry/seed; other categories use 50 (2100 episodes/seed, 6300 over seeds 0--2). Both policies use simulator depth, ten denoising steps, and replanning every 24 actions. We evaluate both policies with three seeds (0--2). \Cref{tab:robodojo_category_seed_results} reports category and overall Score/SR as mean $\pm$ population standard deviation across seeds, together with each seed's overall result; overall metrics average the five categories equally. \Cref{tab:robodojo_per_task} reports per-entry Score/SR averaged over the same three seeds.

\noindent\begin{minipage}{\linewidth}
    \centering
    \captionof{table}{RoboDojo category and seed results. Score (/100) and SR (\%) are mean $\pm$ population standard deviation over seeds 0--2; seed rows give each complete evaluation's category-balanced metrics.}
    \label{tab:robodojo_category_seed_results}
    \small
    \setlength{\tabcolsep}{5pt}
    \renewcommand{\arraystretch}{1.02}
    \begin{tabular*}{\linewidth}{@{\extracolsep{\fill}}lrrrr@{}}
        \toprule
        & \multicolumn{2}{c}{\textbf{PrismWAM}} & \multicolumn{2}{c}{\textbf{PrismWAM$^{*}$}} \\
        \cmidrule(lr){2-3} \cmidrule(lr){4-5}
        Category / seed & Score & SR (\%) & Score & SR (\%) \\
        \midrule
        Generalization & $27.01\pm1.12$ & $19.56\pm0.75$ & $36.71\pm0.93$ & $29.33\pm0.98$ \\
        Precision & $28.70\pm0.98$ & $19.50\pm1.77$ & $39.82\pm2.21$ & $31.08\pm2.66$ \\
        Long-horizon & $37.39\pm1.15$ & $22.08\pm0.77$ & $48.67\pm0.26$ & $31.00\pm0.74$ \\
        Memory & $30.46\pm1.80$ & $30.11\pm1.81$ & $50.96\pm0.20$ & $50.67\pm0.27$ \\
        Open & $1.92\pm0.46$ & $1.50\pm0.54$ & $4.98\pm0.52$ & $4.42\pm0.66$ \\
        \midrule
        \textbf{Overall} & $25.10\pm0.68$ & $18.55\pm0.74$ & $36.23\pm0.55$ & $29.30\pm0.61$ \\
        \midrule
        Seed 0 & 25.21 & 18.58 & 35.58 & 28.70 \\
        Seed 1 & 25.86 & 19.43 & 36.92 & 30.13 \\
        Seed 2 & 24.21 & 17.63 & 36.17 & 29.07 \\
        \bottomrule
    \end{tabular*}
\end{minipage}

\begingroup
\small\setlength{\tabcolsep}{9pt}\renewcommand{\arraystretch}{1.00}
\begin{longtable}{@{}lrrrr@{}}
\caption{All 54 RoboDojo evaluation entries: three-seed mean Score (/100) and SR (\%). Rand. marks randomized Generalization entries.}\label{tab:robodojo_per_task}\label{tab:robodojo_per_task_sr}\label{tab:robodojo_15b_per_task}\\
\toprule
 & \multicolumn{2}{c}{PrismWAM} & \multicolumn{2}{c}{PrismWAM$^{*}$} \\
\cmidrule(lr){2-3}\cmidrule(lr){4-5}
Task & Score & SR & Score & SR \\
\midrule
\endfirsthead
\multicolumn{5}{l}{\small Table \thetable{} (continued)}\\
\toprule
 & \multicolumn{2}{c}{PrismWAM} & \multicolumn{2}{c}{PrismWAM$^{*}$} \\
Task & Score & SR & Score & SR \\
\midrule
\endhead
\midrule\multicolumn{5}{r}{\small Continued on next page}\\\endfoot
\bottomrule\endlastfoot
\multicolumn{5}{@{}l}{\itshape Generalization -- standard}\\*
stack bowls & 80.53 & 77.33 & 88.67 & 86.67 \\
pour liquid into cup & 54.67 & 54.67 & 60.00 & 60.00 \\
push T & 0.00 & 0.00 & 1.33 & 1.33 \\
make toast & 40.67 & 22.67 & 68.67 & 60.00 \\
pack objects into box & 25.73 & 4.00 & 31.33 & 5.33 \\
arrange largest number & 7.80 & 2.67 & 12.47 & 5.33 \\
fold clothes & 62.13 & 58.67 & 78.67 & 77.33 \\
sort nesting dolls by size & 14.67 & 14.67 & 32.00 & 32.00 \\
hang mugs & 42.53 & 20.00 & 73.60 & 61.33 \\
store laptop and headphones & 38.67 & 22.67 & 58.40 & 46.67 \\
sweep blocks & 2.67 & 2.67 & 8.00 & 8.00 \\
stack blocks & 74.60 & 72.00 & 93.20 & 92.00 \\
\multicolumn{5}{@{}l}{\itshape Generalization -- randomized}\\*
stack bowls (Rand.) & 31.87 & 22.67 & 21.20 & 8.00 \\
pour liquid into cup (Rand.) & 45.33 & 45.33 & 36.00 & 36.00 \\
push T (Rand.) & 0.00 & 0.00 & 0.00 & 0.00 \\
make toast (Rand.) & 27.00 & 10.67 & 22.67 & 10.67 \\
pack objects into box (Rand.) & 23.73 & 4.00 & 24.87 & 2.67 \\
arrange largest number (Rand.) & 2.73 & 0.00 & 8.93 & 1.33 \\
fold clothes (Rand.) & 13.07 & 8.00 & 86.40 & 84.00 \\
sort nesting dolls by size (Rand.) & 1.33 & 1.33 & 1.33 & 1.33 \\
hang mugs (Rand.) & 11.27 & 0.00 & 20.33 & 0.00 \\
store laptop and headphones (Rand.) & 40.27 & 22.67 & 32.53 & 17.33 \\
sweep blocks (Rand.) & 0.00 & 0.00 & 5.33 & 5.33 \\
stack blocks (Rand.) & 7.07 & 2.67 & 15.13 & 1.33 \\
\pagebreak
\multicolumn{5}{@{}l}{\itshape Precision}\\*
fasten screws & 32.93 & 8.00 & 58.60 & 30.00 \\
play Xylophone & 4.00 & 4.00 & 0.00 & 0.00 \\
insert tubes & 74.67 & 60.67 & 82.53 & 74.00 \\
deposit coin & 15.33 & 9.33 & 40.00 & 34.67 \\
plug in charger & 1.33 & 1.33 & 10.67 & 10.67 \\
insert key & 14.40 & 0.00 & 14.40 & 0.00 \\
pour balls into vase & 38.67 & 38.67 & 58.00 & 58.00 \\
build tower & 48.27 & 34.00 & 54.33 & 41.33 \\
\multicolumn{5}{@{}l}{\itshape Long-horizon}\\*
put bottles into dustbin & 95.60 & 92.67 & 99.50 & 99.33 \\
fill egg holder & 7.17 & 0.00 & 9.00 & 0.00 \\
play tic tac toe & 40.43 & 10.67 & 31.30 & 4.67 \\
organize table & 59.83 & 7.33 & 55.00 & 4.00 \\
classify objects & 5.70 & 1.33 & 53.20 & 42.67 \\
play stacking toy & 5.00 & 0.00 & 42.93 & 19.33 \\
fill pen holder & 62.03 & 41.33 & 65.73 & 45.33 \\
make kong & 23.33 & 23.33 & 32.67 & 32.67 \\
\multicolumn{5}{@{}l}{\itshape Memory}\\*
cover blocks & 100.00 & 100.00 & 100.00 & 100.00 \\
press by number & 20.00 & 20.00 & 81.33 & 81.33 \\
match and pick from conveyor & 51.33 & 51.33 & 89.33 & 89.33 \\
imitate sorting sequence & 2.77 & 0.67 & 2.40 & 0.67 \\
swap T & 8.67 & 8.67 & 32.67 & 32.67 \\
swap blocks & 0.00 & 0.00 & 0.00 & 0.00 \\
\multicolumn{5}{@{}l}{\itshape Open}\\*
align blocks & 0.00 & 0.00 & 0.00 & 0.00 \\
classify objects by language & 1.07 & 0.00 & 6.00 & 2.67 \\
solve equation & 0.00 & 0.00 & 0.00 & 0.00 \\
pick from conveyor by image & 0.00 & 0.00 & 0.00 & 0.00 \\
stack blocks by language & 7.60 & 5.33 & 11.33 & 10.67 \\
store tools in toolbox & 0.00 & 0.00 & 0.50 & 0.00 \\
general pickup & 6.67 & 6.67 & 22.00 & 22.00 \\
pour by language & 0.00 & 0.00 & 0.00 & 0.00 \\
\midrule
\textbf{Overall (five categories)} & \textbf{25.10} & \textbf{18.55} & \textbf{36.23} & \textbf{29.30} \\
\end{longtable}
\endgroup
\par\endgroup

\end{document}